\documentclass[sigconf,nonacm]{acmart}

\AtBeginDocument{%
  \providecommand\BibTeX{{%
    \normalfont B\kern-0.5em{\scshape i\kern-0.25em b}\kern-0.8em\TeX}}}

\copyrightyear{2023}
\acmYear{2023}
\setcopyright{acmlicensed}\acmConference[KDD '23]{Proceedings of the 29th
ACM SIGKDD Conference on Knowledge Discovery and Data Mining}{August
6--10, 2023}{Long Beach, CA, USA}
\acmBooktitle{Proceedings of the 29th ACM SIGKDD Conference on Knowledge
Discovery and Data Mining (KDD '23), August 6--10, 2023, Long Beach, CA,
USA}
\acmPrice{15.00}
\acmDOI{10.1145/3580305.3599921}
\acmISBN{979-8-4007-0103-0/23/08}

\usepackage{latexsym}
\usepackage{graphics}
\usepackage{array}
\usepackage{booktabs} %
\usepackage{epstopdf}
\usepackage{multirow}
\usepackage{balance}
\usepackage{amsfonts}
\usepackage{amsmath}
\usepackage{graphicx}
\usepackage{subfigure}
\usepackage[noend,ruled]{algorithm2e}
\usepackage{microtype}
\usepackage{xurl} %
\usepackage{enumitem}
\usepackage{tabularx}
\usepackage{caption} %
\usepackage{commath}  %
\usepackage{makecell}  %
\usepackage[hang,flushmargin]{footmisc}  %

\usepackage[T1]{fontenc}

\usepackage[utf8]{inputenc}

\usepackage{microtype}

\def\ddefloop#1{\ifx\ddefloop#1\else\ddef{#1}\expandafter\ddefloop\fi}
\def\ddef#1{\expandafter\def\csname bb#1\endcsname{\ensuremath{\mathbb{#1}}}}
\ddefloop ABCDEFGHIJKLMNOPQRSTUVWXYZ\ddefloop
\def\ddef#1{\expandafter\def\csname bf#1\endcsname{\ensuremath{\mathbf{#1}}}}
\ddefloop ABCDEFGHIJKLMNOPQRSTUVWXYZabcdefghijklmnopqrstuvwxyz\ddefloop
\def\ddef#1{\expandafter\def\csname c#1\endcsname{\ensuremath{\mathcal{#1}}}}
\ddefloop ABCDEFGHIJKLMNOPQRSTUVWXYZ\ddefloop
\def\ddef#1{\expandafter\def\csname v#1\endcsname{\ensuremath{\boldsymbol{#1}}}}
\ddefloop ABCDEFGHIJKLMNOPQRSTUVWXYZabcdefghijklmnopqrstuvwxyz\ddefloop
\def\ddef#1{\expandafter\def\csname v#1\endcsname{\ensuremath{\boldsymbol{\csname #1\endcsname}}}}
\ddefloop {alpha}{beta}{gamma}{delta}{epsilon}{varepsilon}{zeta}{eta}{theta}{vartheta}{iota}{kappa}{lambda}{mu}{nu}{xi}{pi}{varpi}{rho}{varrho}{sigma}{varsigma}{tau}{upsilon}{phi}{varphi}{chi}{psi}{omega}{Gamma}{Delta}{Theta}{Lambda}{Xi}{Pi}{Sigma}{varSigma}{Upsilon}{Phi}{Psi}{Omega}{ell}\ddefloop

\def\E{\mathbb{E}}

\newcommand{\nop}[1]{}

\newcommand{\method}[1]{\mbox{#1}\xspace}

\newcommand{\our}{\method{TwHIN-BERT}}

\DeclareMathOperator*{\argmax}{argmax}

\DeclareMathAlphabet{\mathbbold}{U}{bbold}{m}{n}

\newcommand{\smallsection}[1]{\vspace{1mm}\noindent\textbf{#1.}}  

\begin{document}

\title{TwHIN-BERT: A Socially-Enriched Pre-trained Language Model for Multilingual Tweet Representations at Twitter}

\settopmatter{authorsperrow=4}

\author{Xinyang Zhang}
\authornotemark[1]
\email{xz43@illinois.edu}
\affiliation{%
  \institution{The University of Illinois at Urbana-Champaign}
  \city{Urbana}
  \state{IL}
  \country{USA}
}

\author{Yury Malkov}
\email{ymalkov@twitter.com}
\affiliation{%
  \institution{Twitter Cortex}
  \city{San Francisco}
  \state{CA}
  \country{USA}
}

\author{Omar Florez}
\email{oflorez@twitter.com}
\affiliation{%
  \institution{Twitter Cortex}
  \city{San Francisco}
  \state{CA}
  \country{USA}
}

\author{Serim Park}
\email{serimp@twitter.com}
\affiliation{%
  \institution{Twitter Cortex}
  \city{San Francisco}
  \state{CA}
  \country{USA}
}

\author{Brian McWilliams}
\email{brimcwilliams@twitter.com}
\affiliation{%
  \institution{Twitter Cortex}
  \city{San Francisco}
  \state{CA}
  \country{USA}
}

\author{Jiawei Han}
\email{hanj@illinois.edu}
\affiliation{%
  \institution{The University of Illinois at Urbana-Champaign}
  \city{Urbana}
  \state{IL}
  \country{USA}
}

\author{Ahmed El-Kishky}
\authornotemark[1]
\email{aelkishky@twitter.com}
\affiliation{%
  \institution{Twitter Cortex}
  \city{San Francisco}
  \state{CA}
  \country{USA}
}

\thanks{$^\mathbf{*}$ Corresponding authors: 
xz43@illinois.edu, aelkishky@twitter.com \newline $^1$\url{https://github.com/xinyangz/TwHIN-BERT}}

\renewcommand{\shortauthors}{Xinyang Zhang et al.}

\begin{abstract}

Pre-trained language models (PLMs) are fundamental for natural language processing applications.
Most existing PLMs are not tailored to the noisy user-generated text on social media, and the pre-training does not factor in the valuable social engagement logs available in a social network.
We present \our, a multilingual language model productionized at Twitter, trained on in-domain data from the popular social network. 
\our differs from prior pre-trained language models as it is trained with not only text-based self-supervision but also with a social objective based on the rich social engagements within a Twitter heterogeneous information network (TwHIN).
Our model is trained on $7$ billion tweets covering over $100$ distinct languages, providing a valuable representation to model short, noisy, user-generated text.
We evaluate our model on various multilingual social recommendation and semantic understanding tasks and demonstrate significant metric improvement over established pre-trained language models.
We open-source \our and our curated hashtag prediction and social engagement benchmark datasets to the research community$^1$.
\vspace{-5mm}
\end{abstract}



\maketitle

\section{Introduction}
The proliferation of pre-trained language models (PLMs)~\cite{devlin2019bert,conneau2019unsupervised} based on the Transformer architecture~\cite{vaswani2017attention} has pushed state of the art across many tasks in natural language processing (NLP). As an application of transfer learning, these models are typically trained on massive text corpora and, when fine-tuned on downstream tasks, have demonstrated state-of-the-art performance.

Despite the success of PLMs in general-domain NLP, fewer attempts have been made in language model pre-training for user-generated text on social media.
In this work, we pre-train a language model for Twitter -- a prominent social media platform where users post short messages called Tweets.
Tweets contain informal diction, abbreviations, emojis, and topical tokens such as hashtags. 
As a result, PLMs designed for general text corpora may struggle to understand Tweet semantics accurately.
Existing works~\cite{nguyen2020bertweet, barbieri2022xlmt} on Twitter LM pre-training do not address these challenges and simply replicate general domain pre-training on Twitter corpora.

\begin{figure}[t]
    \centering
    \includegraphics[width=0.88\linewidth]{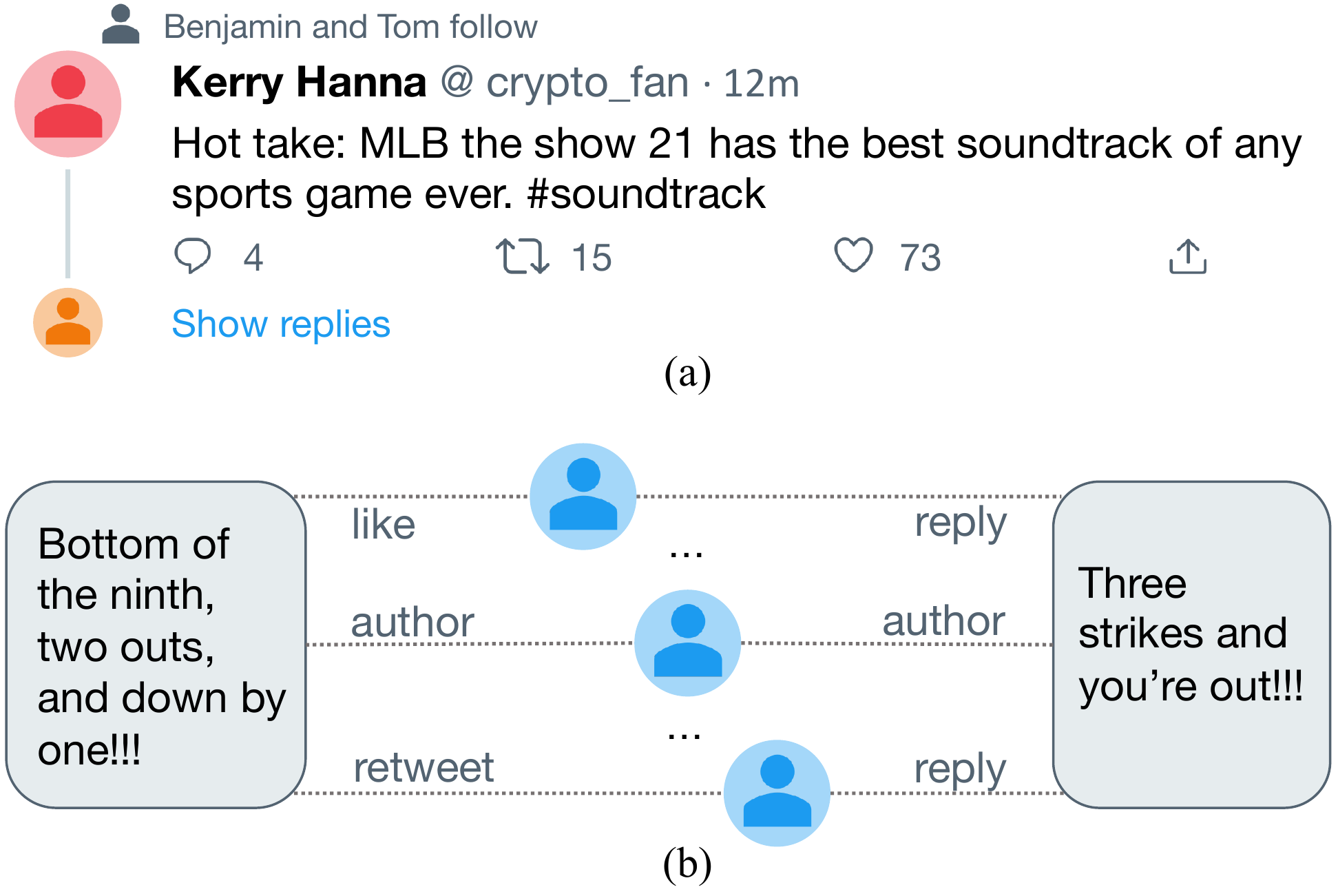}
    \vspace{-3mm}
    \caption{(a) This mock-up shows a short-text Tweet and social engagements such as Faves, Retweets, Replies, Follows that create a social context to Tweets and signify Tweet appeal to engaging users. (b) Co-engagement is a strong indicator of Tweet similarity.}
    \vspace{-5mm}
    \label{fig:tweets}
\end{figure}

A distinctive feature of Twitter social media is the user interactions through Tweet engagements.
As seen in Figure~\ref{fig:tweets}, when a user visits Twitter, in addition to posting Tweets, they can perform a variety of \textit{social} actions such as ``Favoriting'', ``Replying'' and ``Retweeting'' Tweets.
The wealth of such engagement information is invaluable to Tweet content understanding.
For example, the post ``bottom of the ninth, two outs, and down by one!!'' would be connected to baseball topics by its co-engaged Tweets, such as ``three strikes and you're out!!!''.
Without the social contexts, a conventional text-only PLM objective would struggle to build this connection.
As an additional benefit, a socially-enriched language model will also vastly benefit common applications on social media, such as social recommendations~\cite{ying2018pinsage} and information diffusion prediction~\cite{cheng2014cascade,sankar2020infvae}.

We introduce \our -- a multilingual language model for Twitter pre-trained with social engagements.
The key idea of our method is to leverage \emph{socially similar Tweets} for pre-training.
Building on this idea, \our has the following features. (1) We construct a Twitter Heterogeneous Information Network(TwHIN)~\cite{elkishky2022twhin} to unify the multi-typed user engagement logs.
Then, we run scalable embedding and approximate nearest neighbor search to sift through hundreds of billions of engagement records and mine socially similar Tweet pairs.
(2) In conjunction with masked language modeling, we introduce a contrastive social objective that enforces the model to tell if a pair of Tweets are socially similar or not.
Our model is trained on 7 billion Tweets from over 100 distinct languages, of which 1 billion have social engagement logs.

We evaluate the \our model on both social recommendation and semantic understanding downstream evaluation tasks.
To comprehensively evaluate on many languages, we curate two large-scale datasets, a social engagement prediction dataset focused on social aspects and a hashtag prediction dataset focused on language aspects.
In addition to these two curated datasets, we also evaluate on established benchmark datasets to draw direct comparisons to other available pre-trained language models.
\our achieves state-of-the-art performance in our evaluations with a major advantage in the social tasks.

In summary, our contributions are as follows:

\begin{itemize}[leftmargin=*,nosep]
\item We build the first ever socially-enriched pre-trained language model for noisy user-generated text on Twitter.
\item Our model is the strongest multilingual Twitter PLM so far, covering 100 distinct languages.
\item Our model has a major advantage in capturing the social appeal of Tweets.
\item We open-source \our as well as two new Tweet benchmark datasets:  (1) hashtag prediction and (2) social engagement prediction.
\end{itemize}
\begin{figure*}[t]
    \centering
    \includegraphics[width=0.95\linewidth]{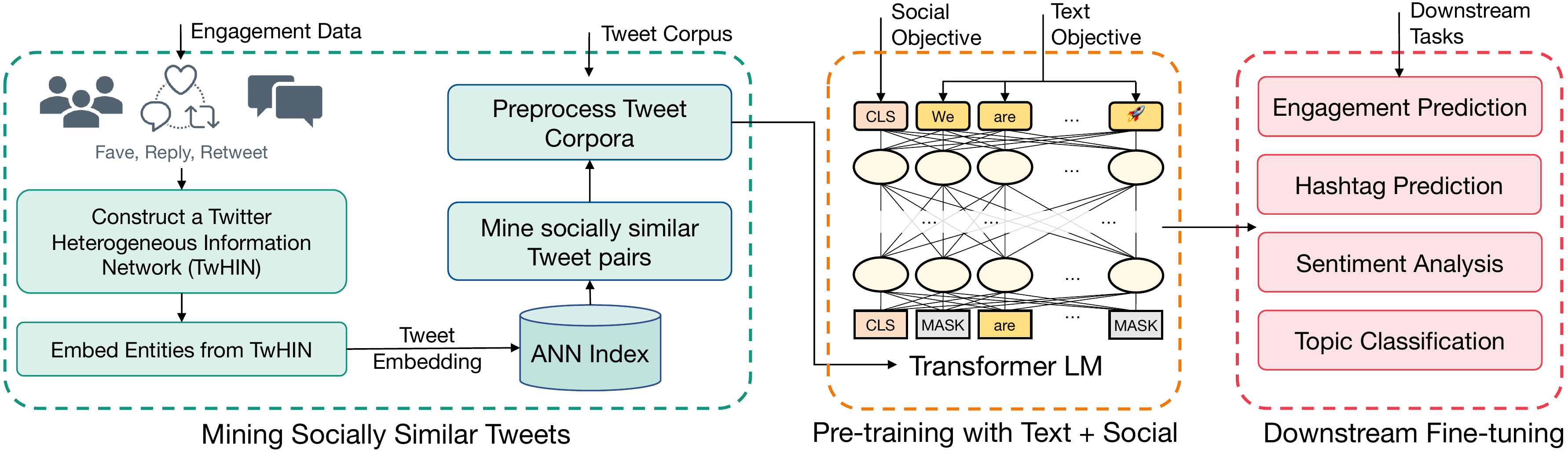}
    \vspace{-3mm}
    \caption{We outline the end-to-end \our process. This three-step process involves (1) mining socially similar Tweet pairs by embedding a Twitter Heterogeneous Information Network  (2) training \our using a joint social and MLM objective and finally (3) fine-tuning \our on downstream tasks.}
    \vspace{-3mm}
    \label{fig:twhin-bert-framework}
\end{figure*}

\section{TwHIN-BERT}
In this section, we outline how we construct training examples for our social pre-training objectives and subsequently train \our with social and text objectives.
As seen in Figure~\ref{fig:twhin-bert-framework}, we first construct and embed a user-Tweet engagement network.
The resultant Tweet embeddings are then used to mine pairs of socially similar Tweets.
These Tweet pairs and others are used to pre-train \our, which can then be fine-tuned for various downstream tasks.

\subsection{Mining Socially Similar Tweets}
With abundant social engagement logs, we (informally) define socially similar Tweets as \emph{Tweets that are co-engaged by a similar set of users}.
The challenge lies in how to implement this social similarity by (1) fusing heterogeneous engagement types, such as ``Favorite'', ``Reply'', ``Retweet'', and (2) efficiently mining billions of similar Tweet pairs.

To address these challenges, \our first constructs a \textbf{Tw}itter \textbf{H}eterogeneous \textbf{I}nformation \textbf{N}etwork (TwHIN) from the engagement logs, then runs a scalable heterogeneous network embedding method to capture co-engagement and map Tweets and users into a vector space.
With this, social similarity translates to embedding space similarity.
Subsequently, we mine similar Tweet pairs via ANN search on the Tweet embeddings.

\subsubsection{Constructing TwHIN}

We define and construct TwHIN as:

\begin{figure}[t]
    \centering
    \includegraphics[width=0.9\linewidth]{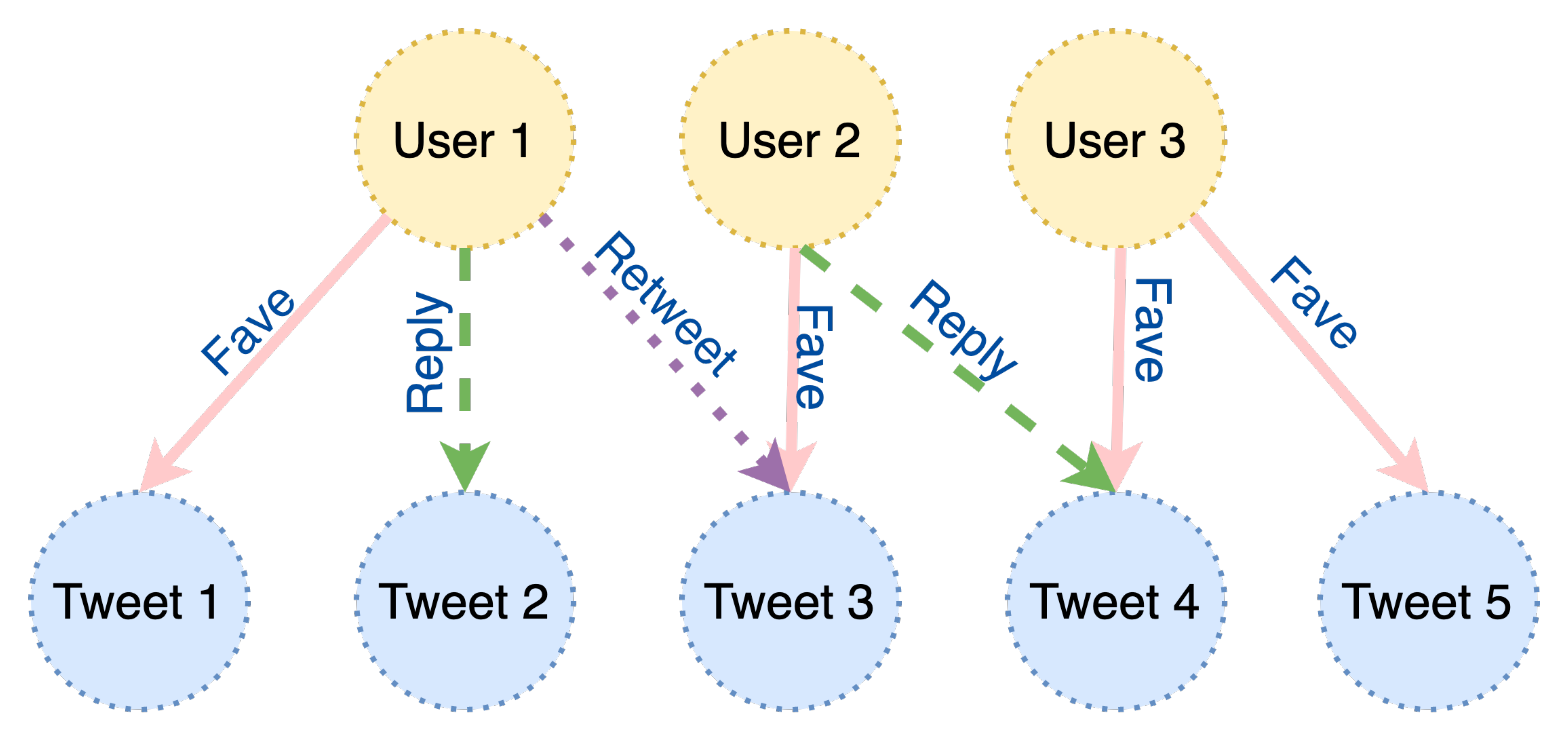}
    \vspace{-3mm}
    \caption{Twitter Heterogeneous Information Network (TwHIN) capturing social engagements between users and Tweets.}
    \vspace{-5mm}
    \label{fig:twhin-bipartite}
\end{figure}

\begin{definition}[TwHIN]
\label{def:twhin}
Our Twitter Heterogeneous Information Network is a directed bipartite graph $G=(U, T, E, \phi)$, where $U$ is the set of user nodes, $T$ is the set of Tweet nodes, $E = U \times T$ is the set of engagement edges.
$\phi: E \mapsto \cR$ is an edge type mapping function.
Each edge $e \in E$ belongs to a type of engagement in $\cR$.
\end{definition}

Our curated TwHIN (Figure~\ref{fig:twhin-bipartite}) consists of approximately 200 million distinct users, 1 billion Tweets, and over 100 billion edges.
We posit that our TwHIN encodes not only user preferences but also Tweet social appeal.
We perform scalable network embedding to derive a social similarity metric from TwHIN.
The network embedding fuses the heterogeneous engagements into a unified vector space that's easy to operate on.

\subsubsection{Embedding TwHIN Nodes}
 We seek to learn shallow embedding vectors (i.e., vector of learnable parameters) for each user ($u_j$) and Tweet ($t_k$) in the TwHIN; we denote these learnable embeddings for users and Tweets as $\mathbf{u_j}$ and $\mathbf{t_k}$
respectively.
While our approach is agnostic to the exact methodology used to embed TwHIN, we follow the approach outlined in~\cite{elkishky2022twhin,el2022knn}. 
A user-Tweet pair for a particular relation type $\phi((u_j, t_k)) = r_m$ is scored with a scoring function of the form $f(u_j, t_k, r_m)$.
Our training objective seeks to learn $\mathbf{u}$, $\mathbf{t}$ and $\mathbf{r}$ parameters that maximize a log-likelihood constructed from the scoring function for $(u, t) \in G$ and minimize for $(u, t) \notin G$.

For simplicity, we apply a simple dot product comparison between user and Tweet representations. For a user-tweet edge $(u_j, t_k)$ of relation $r_m$, this operation is defined by:
\begin{equation}
    \label{eq:scoring}
    f(e) = f(u_j, t_k, r_m) = (\mathbf{u_j} +\mathbf{r_m}) ^\intercal \mathbf{t_k}
\end{equation}

As seen in Equation~\ref{eq:scoring}, we co-embed users and Tweets and scoring is performed by applying an engagement-specific translation embedding to user representations and computing the dot product with a Tweet representation.
The task is then formulated as an edge (or link) prediction task.
Following previous method~\cite{elkishky2022twhin,mikolov2013distributed,goldberg2014word2vec}, we maximize the following negative sampling objective: 
\begin{equation}
    \label{eq:objective}
    \argmax_{\mathbf{u}, \mathbf{r}, \mathbf{t}}\sum_{e \in G} \left[  \log \sigma (f(e)) + \sum_{e' \in N(e)} \log \sigma (-f(e')) \right]
\end{equation}
where $N(e)$ is a set of negatively sampled edges by corrupting positive edges via replacing either the user or Tweet in an edge with a negatively sampled user or Tweet.
As user-Tweet engagement graphs are very sparse, randomly corrupting an edge in the graph is very likely to be a `negative' edge absent from the graph.

Equation~\ref{eq:objective} represents the log-likelihood of predicting a binary ``real" or ``fake'' label for the set of edges in the network (real) along with a set of the ``fake'' negatively sampled edges. 
To maximize the objective, we learn $\mathbf{u}$ and $\mathbf{i}$  parameters to differentiate positive edges from negative, unobserved edges.

We adopt the PyTorch-Biggraph~\cite{lerer2019pytorch} framework for scalability.
Following previous approaches, we train for 10 epochs and perform negative sampling both uniformly and proportional to entity prevalence in TwHIN~\cite{bordes2013translating,lerer2019pytorch}. Optimization is via Adagrad.

Upon learning dense representations of nodes in TwHIN, we utilize the Tweet representations to mine socially similar Tweets.

\subsubsection{Mining Similar Tweet Pairs}
\label{sec:mining_similar_tweets}
Given the learned TwHIN Tweet embeddings, we seek to identify pairs of Tweets with \textit{similar social appeal} -- that is, Tweets that appeal to (i.e., are likely to be engaged with) similar users.
We will use these socially-similar Tweet pairs as self-supervision when training \our.
To identify these pairs, we perform an approximate nearest neighbor (ANN) search in the TwHIN embedding space.
To efficiently perform the search over 1B+ Tweets, we use the optimized FAISS\footnote{\url{https://github.com/facebookresearch/faiss}} toolkit~\cite{johnson2019billion} to create a compact index of Tweets keyed by their engagement-based TwHIN embeddings. 
As each Tweet embedding is 256-dimensional, storing billion-scale Tweet embeddings would require more than one TB of memory.
To reduce the size of the index such that it can fit on a 16 A100 GPU node, with each GPU possessing 40GB of memory, we apply
product quantization~\cite{jegou2010product} to discretize and reduce embeddings size.
The resultant index corresponds to \texttt{OPQ64,IVF65536,PQ64} in the FAISS index factory terminology.

After creating the FAISS index and populating it with TwHIN Tweet embeddings, we search the index using Tweet embedding queries to find pairs of similar Tweets ($t_i, t_j)$ such that $\vt_i$ and $\vt_j$ are close in the embedding space as defined by their cosine distance. To ensure high recall, we query the FAISS index with 2000 probes.  Finally, we select the $k$ closet Tweets with the cosine distance between the query Tweet and retrieved Tweets' embeddings. These pairs are used in our social objective when pre-training \our.

\subsection{Pre-training Objectives}
Given the mined \textit{socially similar} Tweets, we describe our language model training process.
To train \our, we first run the Tweets through the language model and then train the model with a joint contrastive social loss and masked language model loss.

\paragraph{Tweet Encoding with LM.}
We use a Transformer language model to encode each Tweet.
Similar to BERT~\cite{devlin2019bert}, given the tokenized text $\vw_{t} = [w_1, w_2, ..., w_n]$ of a Tweet $t$, we add special tokens to mark the start and end of the Tweet: $\hat{\vw}_t = \text{[CLS]} \vw_t \text{[SEP]}$.
As the Tweets are usually shorter than the maximum sequence length of a language model, we group multiple Tweets and feed them together into the language model when possible.
We then apply \textit{CLS-pooling}, which takes the [CLS] token embedding of each Tweet. These Tweet embeddings are passed through an MLP projection head for the \textit{social loss} computation.

\begin{align}
    [\ve_{t_1}, \ve_{t_2}, ...] &= \text{Pool} \del{\text{LM}([\hat{\vw}_{t_1}, \hat{\vw}_{t_2}, ...])} \\
    \vz_{t} &= \text{MLP}(\ve_t)
\end{align}

\paragraph{Contrastive Social Loss.}
We use a contrastive loss to let our model learn whether two Tweets are socially similar or not. For each batch of $B$ socially similar Tweet pairs $\{(t_i, t_j)\}_B$, we compute the \emph{NT-Xent} loss~\cite{chen2020simclr} with in-batch negatives:
\begin{equation}
    \cL_{\text{social}}(i,j) = -\log \frac{\exp(\text{sim}(\vz_i, \vz_j)) / \tau}{\sum_{\cN_B(i)} \exp(\text{sim}(\vz_i, \vz_k) / \tau)}
\end{equation}
The negatives $\cN_B(i)$ of Tweet $t_i$ are the $(2B-1)$ other Tweets in the batch that are not paired with $t_i$.
We use cosine similarity for function $\text{sim}(\cdot, \cdot)$.
$\tau$ is the loss temperature.

Our overall pre-training objective is a combination of the contrastive social loss and the masked language model loss~\cite{devlin2019bert}:
\begin{equation}
    \cL = \cL_{\text{social}} + \lambda \cL_{\text{MLM}}
\end{equation}
$\lambda$ is a hyperparameter that balances the social and language loss.

\subsection{Pre-training Setup}
\label{sec:pretrain_setup}
\smallsection{Model Architecture}
We use the same Transformer architecture as BERT~\cite{devlin2019bert} for our language model.
We adopt the XLM-R~\cite{conneau2019unsupervised} tokenizer, which offers good capacity and coverage in all languages.
The model has a vocabulary size of 250K.
The max sequence length is set to 128 tokens.
The detailed model setup can be found in Appendix~\ref{appendix:hyperparameters}.
Note that although we have chosen this specific architecture, our social objective can be used in conjunction with a wide range of language model architectures.

\smallsection{Pre-training Data}
We collect 7 billion Tweets in 100 languages from Jan. 2020 to Jun. 2022. Additionally, we collect 100 billion user-Tweet social engagement data covering 1 billion of our Tweets.
We re-sample based on language frequency raised to the power of 0.7 to mitigate the under-representation of low-resource languages.

\smallsection{Training Procedure}
Our training has two stages.
In the first stage, we train the model from scratch using the 6 billion Tweets without user engagement.
The model is trained for 500K steps on 16 Nvidia A100 GPUs (a2-megagpu-16g) with a total batch size of 6K.
In the second stage, the model is trained for another 500K steps on the 1 billion Tweets with the joint MLM and social loss.
We use mixed precision during training.
Overall pre-training takes approximately five days for the base model and two weeks for the large model.
We refer readers to Appendix~\ref{appendix:hyperparameters} for the detailed hyperparameter setup.

\section{Experiments}
In this section, we discuss baseline specifications, evaluation setup, and results from two families of downstream evaluation tasks.

\subsection{Evaluated Methods}
We evaluate \our against the following baselines.

\begin{itemize}[leftmargin=*,nosep]
        \item \textbf{mBERT}~\cite{devlin2019bert}
            is the multilingual language variant of the popular BERT~\cite{devlin2019bert} language model. It is a general domain language model trained on Wikipedia dumps. 
            
        \item \textbf{XLM-R}~\cite{conneau2019unsupervised}
            is a state-of-the-art general domain multilingual language model at its sizes.
            It is trained on over two terabytes of CommonCrawl data.

        \item \textbf{BERTweet}~\cite{nguyen2020bertweet} is the previous state-of-the-art English tweet language model.
        It adopts a monolingual tokenizer trained from scratch on tweets and replicates RoBERTa~\cite{liu2019roberta} training from scratch on 845M English tweets.

        \item \textbf{XLM-T}~\cite{barbieri2022xlmt} is a multilingual Twitter language model based on XLM-R~\cite{conneau2019unsupervised}. It adopts the XLM-R tokenizer and model checkpoint and continues training on over 200M multilingual tweets.

        \item \textbf{\our-MLM} is an ablation of our model. It is trained on the same corpus and with the same protocol as our main models. It uses only an MLM objective.
\end{itemize}

We include \emph{base} and \emph{large} sizes of our model train on the same corpus.
All baselines are \textit{base} variants (with between 135M to 278M parameters depending on the size of the tokenizer). Our large model has around 550M parameters.

We note that all externally published models we compared against were trained on different quantities of data and the data differed temporally and linguistically.
As such, we include these models to demonstrate the gap in performance between widely-used published and open-sourced models and the model we plan on open-sourcing.
On the other hand, our \emph{base-MLM} model draws a direct comparison and isolates the effect of social engagement on the resultant model.

\subsection{Social Engagement Prediction}

\begin{table*}[t]
    \center
    \caption{Engagement prediction HITS@10 on high, mid, low-resource, and average of all languages.}
    \label{tbl:engagement_results}
    \begin{tabular}{l cccc cccc cccc c}
        \toprule
        & \multicolumn{4}{c}{\textbf{High-Resource}} & \multicolumn{4}{c}{\textbf{Mid-Resource}} & \multicolumn{4}{c}{\textbf{Low-Resource}} & \textbf{All} \\
        \cmidrule(lr){2-5} \cmidrule(lr){6-9} \cmidrule(lr){10-13} \cmidrule(lr){14-14}
        \textbf{Method} & \textbf{en} & \textbf{ja} & \textbf{es} & \textbf{ar} & \textbf{el} & \textbf{ur} & \textbf{tl} & \textbf{nl} & \textbf{no} & \textbf{te} & \textbf{da} & \textbf{ps} & \textbf{Avg.} \\
        \midrule
        \textbf{BERTweet} & .1414 & - & - & - & - & - & - & - & - & - & - & - & - \\
        \textbf{mBERT} & .0633 & .0227 & .0575 & .0532 & .0496 & .0437 & .0610 & .0616 & .0731 & .0279 & .1060 & .0522  & .0732 \\
        \textbf{XLM-R} & .0850 & .0947 & .0704 & .0546 & .0628 & .0315 & .0653 & .0650 & .1661 & .0505 & .1150 & .0727  & .0849 \\
        \textbf{XLM-T} & .1181 & .1079 & .1103 & .1403 & .0562 & .0352 & .0877 & .0762 & .1156 & .0728 & .1167 & .0662  & .1043 \\
        \midrule
        \textbf{\our}  \\
        -~\textbf{Base-MLM} & .1400 & .1413 & .1204 & .1640 & .0801 & .0547 & .0700 & .0965 & .1502 & .0883 & .1334 & .0600  & .1161 \\
        -~\textbf{Base} & .1552 & .2065 & .1618 & \textbf{.2206} & .0944 & .0627 & .1030 & \textbf{.1346} & .1920 & .1017 & .1470 & .0799  & .1436 \\
        -~\textbf{Large} & \textbf{.1585} & \textbf{.2325} & \textbf{.2055} & .1989 & \textbf{.1065} & \textbf{.0667} & \textbf{.1053} & .1248 & \textbf{.2118} & \textbf{.1654} & \textbf{.1475} & \textbf{.0817}  & \textbf{.1497} \\
        \bottomrule
    \end{tabular}
\end{table*}

Our first benchmark task is \textit{social engagement prediction}.
This task aims to evaluate how well the pre-trained language models capture the social aspects of user-generated text. In our task, we predict whether users modeled via a user embedding vector will perform a certain social engagement on a given Tweet.

We use different pre-trained language models to generate representations for Tweets, and then feed these representations into a simple prediction model alongside the corresponding user representation.
The engagement prediction model is trained to predict whether a user will engage with a specific Tweet.
The LM-generated embeddings are fixed when we train the downstream engagement prediction model.

\smallsection{Dataset}
To curate our Tweet-Engagement dataset, we select the 50 popular languages on Twitter and sample 10,000
(or all if the total number is less than 10,000) 
Tweets of each language from a fixed time period. All Tweets are available via the Twitter public API.
We then collect the user-Tweet engagement records associated with these Tweets.
There are, on average, 29K engagement records per language.
We ensure that there is no overlap between the evaluation and pre-training datasets. 

Each engagement record consists of a pre-trained 256-dimensional user embedding~\cite{elkishky2022twhin} and a Tweet ID that indicates the user has engaged with the given Tweet.
To ensure privacy, each user embedding appears only once, however, each tweet may be engaged by multiple users.
We split the Tweets into train, development, and test sets with a 0.8/0.1/0.1 ratio, and then collect the respective engagement records for each subset.

\smallsection{Prediction Model}
Given a pre-trained language model, we use it to generate an embedding for each Tweet $t$ given its content $\vw_t$: $\ve_{t} = \text{Pool}\del{\text{LM}(\vw_t)}$.

We apply the following pooling strategies to calculate the Tweet embedding from the language model.
First, we take [CLS] token embedding as the first part.
Then, we take the average token embedding of non-special tokens as the second part.
The two parts are concatenated to form the \emph{Combined} embedding of a Tweet.

With LM-derived Tweet embeddings, pre-trained user embeddings, and the user-Tweet engagement records, we build an engagement prediction model $\vTheta = (\vW_t, \vW_u)$.
Given a user $u$ and a Tweet $t$, the model projects the user embedding $\ve_u$ and the Tweet embedding $\ve_t$ into the same space, and then calculates the probability of engagement:

\begin{align*}
    &\vh_u = \vW_u^{T}\ve_u, \quad \vh_t = \vW_t^{T}\ve_t \\
    &P(t \mid u) = \sigma\del{\vh_u^{T}\vh_t}
\end{align*}

We optimize a negative sampling loss on the training engagement records $R$.
For each engagement pair $(u, t) \in R$, the loss is:
\[
    \log \sigma\del{\vh_u^{T}\vh_t} + \E_{t'\sim P_n(R)}\log \sigma\del{-\vh_u^T \vh_{t'}}
\]
where $P_n(R)$ is a negative sampling distribution. We use the frequency of each Tweet in $R$ to the power of $3/4$ for this distribution.

Our prediction model closely resembles classical link prediction models such as ~\cite{tang2015line}.
We keep the model simple, making sure it will not overpower the language model embeddings.

\smallsection{Evaluation Setup and Metrics}
We conduct a hyperparameter search on the English development dataset and use these hyperparameters for the other languages.
The prediction model projects user and Tweet embedding to 128 dimensions.
We set the batch size to 512, and the learning rate to 1e-3.
The best model on the validation set is selected for test set evaluation.

In the test set, we pair each user with 1,000 Tweets: one Tweet they have engaged with, and the rest are randomly sampled negatives. The model ranks the Tweets by the predicted probability of engagement, and we evaluate with HITS@10. We report median results from 6 runs with different initialization.

\smallsection{Results}
We show summarized results for selected high, mid, and low-resource languages (determined by language frequency on Twitter) in Table~\ref{tbl:engagement_results}.
Language abbreviations are ISO language codes\footnote{\url{https://www.iso.org/iso-639-language-codes.html}}.
We also show the average results from all 50 languages in the evaluation dataset and leave the details in Table~\ref{tbl:engagement_all_lang}.
Our \our model demonstrates significant improvement over the baselines on the social engagement task.
Comparing our model to the ablation without the social loss, we can see the contrastive social pre-training provides a significant lift over just MLM pre-training for social engagement prediction.
An analysis of all 50 evaluation languages shows the \emph{large} model to perform better than the \emph{base} model on average, with more wins than losses.
Additionally, we also observe that our method yields the most improvement when using the \emph{Combined} [CLS] token and average non-special token embedding.
We believe the [CLS] token embedding from our model captures the social aspects of the Tweet while averaging the other token embeddings captures the semantic aspects of the Tweet. Naturally, utilizing both aspects is essential to better model a Tweet's appeal and a user's inclination to engage with a Tweet.

\begin{table}[t]
    \center
    \caption{Text classification dataset statistics. $^*$Statistics for Hashtag shows the numbers for each language.}
    \vspace{-3mm}
    \label{tbl:text_classification_stats}
        \begin{tabular}{l ccccc}
            \toprule
            \textbf{Dataset} & Lang. & Label & Train & Dev & Test\\
            \midrule
            \textbf{SE2017} & en & 3 & 45,389 & 2,000 & 11,906 \\
            \textbf{SE2018-en} & en & 20 & 45,000 & 5,000 & 50,000 \\
            \textbf{SE2018-es} & es & 19 & 96,142 & 2,726 & 9,969 \\
            \textbf{ASAD} & ar & 3 & 137,432 & 15,153 & 16,842 \\
            \textbf{COVID-JA} & ja & 6 & 147,806 & 16,394 & 16,394 \\
            \textbf{SE2020-hi} & hi+en & 3 & 14,000 & 3,000 & 3,000 \\
            \textbf{SE2020-es} & es+en & 3 & 10,800 & 1,200 & 3,000 \\
            \textbf{Hashtag} & multi & 500$^*$ & 16,000$^*$ & 2,000$^*$ & 2,000$^*$ \\
            \bottomrule
        \end{tabular}
    \vspace{-3mm}
\end{table}

\begin{table*}[t]
    \center
    \caption{Multilingual hashtag prediction Macro-F1 on high, mid, low resource, and average of all languages.}
    \vspace{-3mm}
    \label{tbl:hashtag_results}
    \begin{tabular}{l cccc cccc cccc c}
        \toprule
        & \multicolumn{4}{c}{\textbf{High-Resource}} & \multicolumn{4}{c}{\textbf{Mid-Resource}} & \multicolumn{4}{c}{\textbf{Low-Resource}} & \textbf{All} \\
        \cmidrule(lr){2-5} \cmidrule(lr){6-9} \cmidrule(lr){10-13} \cmidrule(lr){14-14}
        \textbf{Method} & \textbf{en} & \textbf{ja} & \textbf{es} & \textbf{ar} & \textbf{el} & \textbf{ur} & \textbf{tl} & \textbf{nl} & \textbf{no} & \textbf{te} & \textbf{da} & \textbf{ps} & \textbf{Avg.} \\
        \midrule
        \textbf{BERTweet} & 59.01 & - & - & - & - & - & - & - & - & - & - & -  & - \\
\textbf{mBERT} & 54.56 & 68.43 & 42.48 & 38.48 & 44.00 & 36.44 & 52.96 & 39.75 & 46.09 & 49.54 & 59.54 & 29.41  & 50.05 \\
\textbf{XLM-R} & 53.90 & 69.07 & 43.80 & 37.85 & 43.94 & 37.56 & 52.99 & 40.85 & 48.94 & 51.47 & 60.35 & 34.92  & 50.86 \\
\textbf{XLM-T} & 55.08 & 70.55 & 45.85 & 42.27 & 44.15 & 39.22 & 54.86 & 41.01 & 49.22 & 52.45 & 59.97 & 33.27  & 51.74 \\
        \midrule     
        \textbf{\our} \\
        \textbf{-~Base-MLM} & 58.38 & 72.66 & 48.41 & 43.08 & 46.89 & 41.53 & 56.76 & 42.36 & 49.60 & 51.13 & 61.00 & 35.37  & 53.66 \\
        \textbf{-~Base} & 59.31 & \textbf{73.03} & 48.59 & 44.24 & \textbf{47.59} & 42.81 & 57.33 & 42.69 & 51.11 & 56.66 & 60.33 & 36.21  & 54.62 \\
        \textbf{-~Large} & \textbf{60.07} & 72.91 & \textbf{49.88} & \textbf{45.41} & 47.43 & \textbf{43.39} & \textbf{59.43} & \textbf{44.80} & \textbf{51.34} & \textbf{57.03} & \textbf{61.56} & \textbf{38.24}  & \textbf{55.23} \\
        \bottomrule
    \end{tabular}
\end{table*}

\begin{table*}[t]
    \center
    \caption{External classification benchmark results.}
    \vspace{-3mm}
    \label{tbl:cls_results}
    \begin{tabular}{l cccccccc}
        \toprule
        & \textbf{SE2017} & \multicolumn{2}{c}{\textbf{SE2018}} & \textbf{ASAD} & \textbf{COVID-JA} & \multicolumn{2}{c}{\textbf{SE2020}} & \textbf{Avg.} \\
        \textbf{Method} & \textbf{en} & \textbf{en} & \textbf{es} & \textbf{ar} & \textbf{ja} & \textbf{hi+en} & \textbf{es+en} & \\
        \midrule
        \textbf{BERTweet} & 72.97 & 33.27 & - & - & - & - & -  & - \\
\textbf{mBERT} & 66.17 & 27.73 & 19.19 & 69.08 & 80.57 & 66.55 & 45.31  & 53.51 \\
\textbf{XLM-R} & 71.15 & 30.94 & 21.05 & 79.09 & 81.67 & 69.59 & 48.97  & 57.49 \\
\textbf{XLM-T} & 72.01 & 31.97 & 21.49 & 80.70 & 81.48 & 70.94 & 51.06  & 58.52 \\
        \midrule
        \textbf{\our} \\
        -~\textbf{Base-MLM} & 72.10 & 32.44 & 21.79 & 80.48 & 82.12 & 72.42 & 51.67  & 59.00 \\
        -~\textbf{Base} & 72.30 & 32.41 & 22.23 & 80.73 & 82.37 & 71.30 & 54.32  & 59.38 \\
        -~\textbf{Large} & \textbf{73.10} & \textbf{33.31} & \textbf{22.80} & \textbf{81.19} & \textbf{82.50} & \textbf{73.08} & \textbf{54.47}  & \textbf{60.06} \\
        \bottomrule
    \end{tabular}
\end{table*}

\subsection{Tweet Classification}
\label{sec:classification}
Our second collection of downstream tasks is Tweet classification. In these tasks, we take as input the Tweet text and predict discrete labels corresponding to the label space for each task.

\smallsection{Datasets}
We curate a multilingual Tweet hashtag prediction dataset (available via Twitter public API) to comprehensively cover the popular languages on Twitter.
In addition, we evaluate on five external benchmark datasets for tasks such as sentiment classification, emoji prediction, and topic classification in selected languages.
We show the dataset statistics in Table~\ref{tbl:text_classification_stats}.

\begin{itemize}[leftmargin=*,nosep]
    \item \textbf{Tweet Hashtag Prediction} dataset is a multilingual hashtag prediction dataset we collected from Tweets. It contains Tweets from 50 popular languages. For each language, the 500 most popular hashtags were selected, and 100k tweets with those hashtags were sampled. We ensured each Tweet will only contain one of the 500 candidate hashtags. Similar to the work proposed in \citet{mireshghallah2022}, the task is to predict the hashtag used in the Tweet.
    \item \textbf{SemEval2017} task 4A~\cite{semeval2017} is a English Tweet sentiment analysis dataset. The labels are three-point sentiments of ``positive'', ``negative'', ``neutral''. 
    \item \textbf{ASAD}~\cite{asad} is an Arabic Tweet sentiment  dataset with the same three-point labels as SemEval2017 T4A.
    \item \textbf{SemEval2020} task 9~\cite{semeval2020t9} contains code-mixed Tweets of Hindi + English and Spanish + English. We use the three-point sentiment analysis part of the dataset for evaluation.
    \item \textbf{SemEval2018} task 2~\cite{semeval2018t2} is an emoji prediction dataset in both English and Spanish. The objective is to predict the most likely used emoji in a Tweet.
    \item \textbf{COVID-JA}~\cite{covid_jp} is a Japanese Tweets classification dataset. The objective is to classify each Tweet into one of the six pre-defined topics around COVID-19.

\end{itemize}

\smallsection{Setup and Evaluation Metrics}
We use the standard language model fine-tuning method as described in ~\cite{devlin2019bert}  and apply a linear prediction layer on top of the pooled output of the last transformer layer.
Each model is fine-tuned for up to 30 epochs, and we evaluate the best model from the training epochs on the test set based on the development set performance.
The fine-tuning hyperparameter setup can be found in Appendix~\ref{appendix:hyperparameters}.
We report the median results from 3 fine-tuning runs with different random seeds.
Results are the evaluation metrics recommended for each benchmark dataset or challenge (Appendix~\ref{appendix:metrics}).
For hashtag prediction datasets, we report macro-F1 scores.
We conduct data contamination tests using character-level 50-gram overlaps~\cite{radford2019gpt2} and found 1.56\% and 1.10\% contamination in the average results reported in Table~\ref{tbl:hashtag_results} and Table~\ref{tbl:cls_results}.

\smallsection{Multilingual Hashtag Prediction}
In Table~\ref{tbl:hashtag_results}, we show macro F1 scores on selected languages from our multilingual hashtag prediction dataset.
We also report the average performance of all 50 languages in the dataset, and leave detailed results in Table~\ref{tbl:hashtag_all_lang}.
We can see that \our significantly outperforms the baseline methods at the same \emph{base} size.
Our \emph{large} model is slightly better than or on par with the \emph{base} model, with a better overall performance.
On the English dataset, our model outperforms the BERTweet monolingual language model trained exclusively on English Tweets and with a dedicated English tokenizer.
Comparing our model to the ablation with no social loss, the two models demonstrate similar performance with our model being slightly better.
These results show that while our model has a major advantage on social tasks, it retains high performance on semantic understanding applications.

\smallsection{External Classification Benchmarks}
As shown in Table~\ref{tbl:cls_results}, our \our matches or outperforms the multilingual baselines on the established classification benchmarks.
BERTweet fares better than our \emph{base} model with its dedicated large English tokenizer and monolingual training.
Our \emph{large} model outperforms all the baselines.
We note that it is not uncommon for a monolingual PLM to perform better than its multilingual counterpart, as observed in~\cite{conneau2019xlm,rust2021multilingual,xue2021mt5}.
Similar to hashtag prediction, \our performs on par with or slightly better than the MLM-only ablation.

\subsection{Varying Downstream Supervision}
In this set of experiments, we study how \our performs when the amount of downstream supervision changes.
We fine-tune our model and baseline models on the hashtag prediction dataset (Section~\ref{sec:classification}).
We select English and Japanese as they are the most popular languages on Twitter.
We change the number of training examples given to the models during fine-tuning.
It is varied from 2 to 32 labeled training examples per class.
We follow the same protocols as Section~\ref{sec:classification} and report macro F1 scores on the test set. 

\begin{figure}[t]
    \centering
    \includegraphics[width=0.95\linewidth]{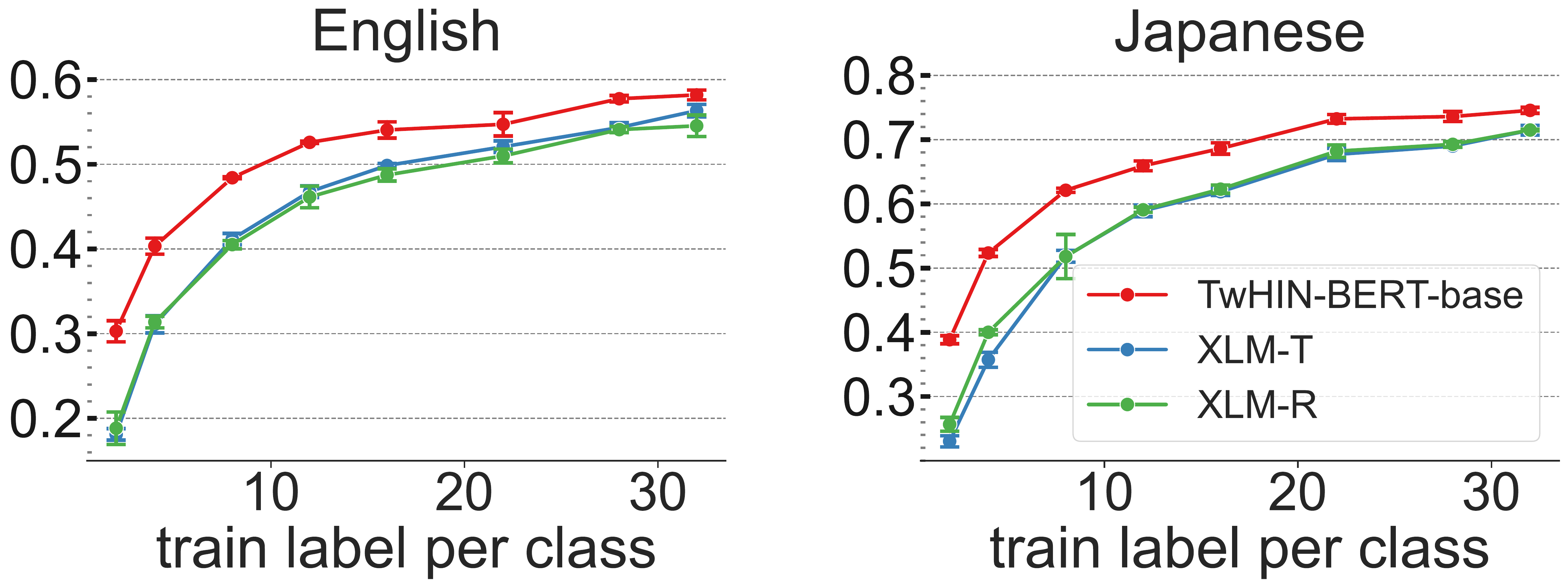}
    \vspace{-3mm}
    \caption{Macro-F1 on English and Japanese hashtag prediction datasets w.r.t. the number of labeled training examples per class.}
    \vspace{-3mm}
    \label{fig:data_ablation}
\end{figure}

Figure~\ref{fig:data_ablation} shows the results.
\our holds significant performance gain across different amount of downstream supervision.
Note that when supervision is scarce, e.g., two labeled training examples per class given, our model has an even larger relative performance improvement over the baselines.
The results indicate that our model may empower weakly supervised applications on Tweet natural language understanding.

\subsection{Feature-based Classification}
\label{sec:appendix}

In addition to language model fine-tuning experiments, we evaluate \our's performance as a feature extractor.
We use the hashtag prediction datasets (Section~\ref{sec:classification}) and select three popular languages with different scripts. We use our model and the baseline models to embed each Tweet into a feature vector and train a Logistic Regression classifier with the fixed feature vectors as input.

Table~\ref{tbl:feature_classification} shows \our outperforming the baselines with a wide margin on all languages. This not only shows \our has learned superior Tweet representations but also showcases its potential in other feature-based downstream applications.

\begin{table}[t]
    \center
    \caption{Feature-based classification on hashtag prediction datasets (Macro-F1).}
    \vspace{-3mm}
    \label{tbl:feature_classification}
    \setlength\tabcolsep{8pt}
    \begin{tabular}{l ccc}
        \toprule
        \textbf{Method} & \textbf{English} & \textbf{Japanese} & \textbf{Arabic} \\
        \midrule
        \textbf{BERTweet} & 48.56 & - & - \\
        \textbf{XLM-R} & 30.88 & 41.14 & 21.55 \\
        \textbf{XLM-T} & 41.66 & 51.56 & 32.46 \\
        \midrule
\textbf{\our-base} & 51.16 & \textbf{64.12} & 37.20 \\
\textbf{\our-large} & \textbf{54.12} & 64.03 & \textbf{38.78} \\
        \bottomrule
    \end{tabular}
    \vspace{-3mm}
\end{table}

\begin{table*}
    \begin{minipage}{\columnwidth}
        \centering
        \caption{Full social engagement prediction results (HITS@10) on all evaluation Languages.}
        \label{tbl:engagement_all_lang}
        \setlength\extrarowheight{0.8pt}
        \setlength\tabcolsep{3pt}
        \scalebox{0.87}{
        \begin{tabular}{l cccccc}
            \toprule
            & & & & \multicolumn{3}{c}{\textbf{\our}} \\
            \cmidrule(lr){5-7}
            \textbf{Language} & \textbf{mBERT} & \textbf{XLM-R} & \textbf{XLM-T} & \textbf{Base-MLM} & \textbf{Base} & \textbf{Large} \\
            \midrule
            \textbf{English (en)} & .0633 & .0850 & .1181 & .1400 & .1552 & \textbf{.1585}  \\
            \textbf{Japanese (ja)} & .0227 & .0947 & .1079 & .1413 & .2065 & \textbf{.2325}  \\
            \textbf{Turkish (tr)} & .0348 & .0476 & .1180 & \textbf{.1268} & .1204 & .0547  \\
            \textbf{Spanish (es)} & .0575 & .0704 & .1103 & .1204 & .1618 & \textbf{.2055}  \\
            \textbf{Arabic (ar)} & .0532 & .0546 & .1403 & .1640 & \textbf{.2206} & .1989  \\
            \textbf{Portuguese (pt)} & .0731 & .1285 & .1709 & .1201 & \textbf{.1924} & .1915  \\
            \textbf{Persian (fa)} & .0556 & .1621 & .1754 & .1903 & .2065 & \textbf{.2097}  \\
            \textbf{Korean (ko)} & .0275 & .1105 & .1446 & .1675 & .3611 & \textbf{.3714}  \\
            \textbf{French (fr)} & .0488 & .0635 & .0805 & .0700 & .1030 & \textbf{.1053}  \\
            \textbf{Russian (ru)} & .0889 & .1482 & .1530 & .0990 & \textbf{.1726} & .1704  \\
            \textbf{German (de)} & .0852 & .1071 & .3019 & .2189 & \textbf{.3020} & .2621  \\
            \textbf{Thai (th)} & .0659 & .1027 & .1056 & .1196 & \textbf{.2083} & .2004  \\
            \textbf{Italian (it)} & .0586 & .0769 & .1237 & .1478 & .1699 & \textbf{.1706}  \\
            \textbf{Hindi (hi)} & .0870 & .0838 & .1140 & .1054 & .1737 & \textbf{.1751}  \\
            \textbf{Indonesian (id)} & .0809 & .0735 & .0921 & .1014 & .1021 & \textbf{.1115}  \\
            \textbf{Polish (pl)} & .0867 & .0835 & .1031 & .1402 & \textbf{.1696} & .1633  \\
            \textbf{Urdu (ur)} & .0437 & .0315 & .0352 & .0547 & .0627 & \textbf{.0667}  \\
            \textbf{Filipino (tl)} & .0610 & .0653 & .0877 & .1045 & .1332 & \textbf{.1400}  \\
            \textbf{Egpt. Arabic (arz)} & .0669 & .0749 & .1049 & .0943 & \textbf{.1159} & .1122  \\
            \textbf{Greek (el)} & .0496 & .0628 & .0562 & .0801 & .0944 & \textbf{.1065}  \\
            \textbf{Serbian (sr)} & .1013 & .1144 & .1359 & .1394 & \textbf{.1647} & .1556  \\
            \textbf{Dutch (nl)} & .0616 & .0650 & .0762 & .0965 & \textbf{.1346} & .1248  \\
            \textbf{Hebrew (he)} & .0392 & .0433 & .0441 & .0499 & .0550 & \textbf{.0577}  \\
            \textbf{Ukrainian (uk)} & .0497 & \textbf{.0842} & .0669 & .0711 & .0811 & \textbf{.0842}  \\
            \textbf{Catalan (ca)} & .1339 & .1364 & .1650 & .1930 & \textbf{.1955} & .1713  \\
            \textbf{Swedish (sv)} & .0942 & .0716 & .1161 & .1342 & \textbf{.1467} & .1462  \\
            \textbf{Tamil (ta)} & .0556 & .0691 & .0929 & .1005 & .1037 & \textbf{.1060}  \\
            \textbf{Finnish (fi)} & .0876 & .1067 & .1317 & .1529 & .1710 & \textbf{.1809}  \\
            \textbf{Czech (cs)} & .1155 & .0904 & .0766 & .0997 & .1062 & \textbf{.1308}  \\
            \textbf{Nepali (ne)} & .0421 & .0555 & .0486 & .0589 & .0787 & \textbf{.0851}  \\
            \textbf{Azerbaijani (az)} & .1561 & .1148 & .1702 & .1576 & .1712 & \textbf{.1839}  \\
            \textbf{Marathi (mr)} & .0506 & .0600 & .0519 & .0597 & .0780 & \textbf{.0906}  \\
            \textbf{Bangla (bn)} & .1361 & .1350 & .1320 & .1601 & .1649 & \textbf{.1675}  \\
            \textbf{Norwegian (no)} & .0731 & .1661 & .1156 & .1502 & .1920 & \textbf{.2118}  \\
            \textbf{Telugu (te)} & .0279 & .0505 & .0728 & .0883 & .1017 & \textbf{.1654}  \\
            \textbf{Pashto (ps)} & .0522 & .0727 & .0662 & .0600 & .0799 & \textbf{.0817}  \\
            \textbf{Danish (da)} & .1060 & .1150 & .1167 & .1334 & .1470 & \textbf{.1475}  \\
            \textbf{Vietnamese (vi)} & .0929 & .1060 & .1085 & .1216 & .1417 & \textbf{.1809}  \\
            \textbf{Cen. Kurdish (ckb)} & .0725 & .0699 & .0946 & .1023 & .1023 & \textbf{.1185}  \\
            \textbf{Gujarati (gu)} & .0666 & .0676 & .0676 & .0793 & .1054 & \textbf{.1057}  \\
            \textbf{Macedonian (mk)} & .0685 & .0945 & .0534 & .0973 & \textbf{.1089} & .1041  \\
            \textbf{Cebuano (ceb)} & .1222 & .1267 & .1767 & .1900 & .2003 & \textbf{.2334}  \\
            \textbf{Romanian (ro)} & .1718 & .1493 & .1991 & .2071 & \textbf{.2264} & \textbf{.2264}  \\
            \textbf{Kannada (kn)} & .0552 & .1355 & .0814 & .1098 & .1282 & \textbf{.2113}  \\
            \textbf{Latvian (lv)} & .0480 & .0297 & .0493 & .0642 & .0655 & \textbf{.0750}  \\
            \textbf{Bulgarian (bg)} & .1953 & .0448 & .0702 & .1790 & .2248 & \textbf{.2269}  \\
            \textbf{Sinhala (si)} & .0504 & .0142 & .0378 & .0630 & \textbf{.0709} & .0661  \\
            \textbf{Icelandic (is)} & .0319 & .0341 & .0466 & .0364 & .0387 & \textbf{.0603}  \\
            \textbf{Sindhi (sd)} & .0619 & .0288 & .0553 & .0885 & .0951 & \textbf{.0973}  \\
            \textbf{Amharic (am)} & .0293 & .0663 & .0491 & .0543 & .0698 & \textbf{.0818}  \\
            \midrule
            \textbf{Average} & .0732 & .0849 & .1043 & .1161 & .1436 & \textbf{.1497}  \\
            \bottomrule
        \end{tabular}
        }
    \end{minipage}\hfill %
    \begin{minipage}{\columnwidth}
        \centering
        \caption{Full hashtag prediction results (Macro-F1) on all evaluation languages.}
        \label{tbl:hashtag_all_lang}
        \setlength\tabcolsep{2.5pt}
        \setlength\extrarowheight{0.8pt}
        \scalebox{0.87}{
        \begin{tabular}{l cccccc}
            \toprule
            & & & & \multicolumn{3}{c}{\textbf{\our}} \\
            \cmidrule(lr){5-7}
            \textbf{Language} & \textbf{mBERT} & \textbf{XLM-R} & \textbf{XLM-T} & \textbf{Base-MLM} & \textbf{Base} & \textbf{Large} \\
            \midrule
            \textbf{English (en)} & 54.56 & 53.90 & 55.08 & 58.38 & 59.31 & \textbf{60.07} \\
            \textbf{Japanese (ja)} & 68.43 & 69.07 & 70.55 & 72.66 & \textbf{73.03} & 72.91 \\
            \textbf{Turkish (tr)} & 42.87 & 46.37 & 47.14 & 48.72 & 49.31 & \textbf{51.12} \\
            \textbf{Spanish (es)} & 42.48 & 43.80 & 45.85 & 48.41 & 48.59 & \textbf{49.88} \\
            \textbf{Arabic (ar)} & 38.48 & 37.85 & 42.27 & 43.08 & 44.24 & \textbf{45.41} \\
            \textbf{Portuguese (pt)} & 47.81 & 50.33 & 51.98 & 52.15 & 52.98 & \textbf{56.08} \\
            \textbf{Persian (fa)} & 43.39 & 45.04 & 45.25 & 46.02 & 47.46 & \textbf{47.94} \\
            \textbf{Korean (ko)} & 75.46 & 77.73 & 78.45 & 79.49 & 79.11 & \textbf{80.02} \\
            \textbf{French (fr)} & 40.37 & 40.81 & 41.89 & 44.43 & 45.40 & \textbf{47.01} \\
            \textbf{German (de)} & 40.80 & 41.42 & 41.11 & 41.32 & 41.38 & \textbf{42.59} \\
            \textbf{Thai (th)} & 44.10 & 56.27 & 57.40 & 58.25 & 58.80 & \textbf{59.46} \\
            \textbf{Italian (it)} & 42.36 & 41.82 & 42.76 & 45.11 & 44.18 & \textbf{45.72} \\
            \textbf{Hindi (hi)} & 49.84 & 51.92 & 52.58 & 55.17 & 55.28 & \textbf{57.29} \\
            \textbf{Chinese (zh)} & 72.88 & 72.54 & 72.40 & 73.85 & \textbf{73.94} & 72.30 \\
            \textbf{Polish (pl)} & 48.97 & 50.20 & 50.50 & 51.20 & 51.81 & \textbf{54.49} \\
            \textbf{Urdu (ur)} & 36.44 & 37.56 & 39.22 & 41.53 & 42.81 & \textbf{43.39} \\
            \textbf{Filipino (tl)} & 52.96 & 52.99 & 54.86 & 56.76 & 57.33 & \textbf{59.43} \\
            \textbf{Greek (el)} & 44.00 & 43.94 & 44.15 & 46.89 & \textbf{47.59} & 47.43 \\
            \textbf{Serbian (sr)} & 42.50 & 42.32 & 40.71 & 44.22 & 45.95 & \textbf{47.45} \\
            \textbf{Dutch (nl)} & 39.75 & 40.85 & 41.01 & 42.36 & 42.69 & \textbf{44.80} \\
            \textbf{Catalan (ca)} & 48.61 & 47.85 & 48.79 & 51.72 & 52.60 & \textbf{52.90} \\
            \textbf{Swedish (sv)} & 47.79 & 47.80 & 47.31 & 49.39 & 51.28 & \textbf{51.44} \\
            \textbf{Tamil (ta)} & 48.04 & 49.67 & 50.65 & 52.85 & 54.14 & \textbf{54.92} \\
            \textbf{Finnish (fi)} & 45.28 & 45.28 & 44.03 & 43.98 & 45.59 & \textbf{46.42} \\
            \textbf{Czech (cs)} & 53.03 & 52.60 & 52.89 & 55.01 & 55.93 & \textbf{56.02} \\
            \textbf{Nepali (ne)} & 44.58 & 47.00 & 46.94 & 49.83 & \textbf{51.57} & 51.12 \\
            \textbf{Marathi (mr)} & 50.85 & 48.40 & 51.44 & 54.18 & \textbf{55.76} & 55.31 \\
            \textbf{Malayalam (ml)} & 38.43 & 42.20 & 42.77 & 44.72 & \textbf{45.86} & 44.36 \\
            \textbf{Bangla (bn)} & 57.79 & 57.08 & 56.74 & 59.11 & 60.32 & \textbf{60.92} \\
            \textbf{Hungarian (hu)} & 60.29 & 59.94 & 60.08 & 61.86 & \textbf{63.81} & 62.68 \\
            \textbf{Slovenian (sl)} & 58.79 & 59.68 & 59.13 & 61.18 & 62.34 & \textbf{62.74} \\
            \textbf{Norwegian (no)} & 46.09 & 48.94 & 49.22 & 49.60 & 51.11 & \textbf{51.34} \\
            \textbf{Telugu (te)} & 49.54 & 51.47 & 52.45 & 55.13 & 56.66 & \textbf{57.03} \\
            \textbf{Pashto (ps)} & 29.41 & 34.92 & 33.27 & 35.37 & 36.21 & \textbf{38.24} \\
            \textbf{Danish (da)} & 59.54 & 60.35 & 59.97 & 61.00 & 60.33 & \textbf{61.56} \\
            \textbf{Cen. Kurdish (ckb)} & 40.28 & 37.59 & 39.06 & 42.89 & \textbf{45.65} & 45.26 \\
            \textbf{Gujarati (gu)} & 52.55 & 54.09 & 54.24 & 57.36 & \textbf{58.59} & 58.54 \\
            \textbf{Romanian (ro)} & 71.24 & 71.53 & 72.34 & 73.17 & 73.25 & \textbf{73.58} \\
            \textbf{Kannada (kn)} & 54.19 & 55.76 & 56.68 & 59.09 & \textbf{61.34} & 60.19 \\
            \textbf{Estonian (et)} & 57.81 & 58.10 & 59.00 & 61.95 & 61.22 & \textbf{62.61} \\
            \textbf{Latvian (lv)} & 58.03 & 55.18 & 57.53 & 58.43 & 59.52 & \textbf{61.47} \\
            \textbf{Bulgarian (bg)} & 65.20 & 65.52 & 66.45 & 67.42 & 66.94 & \textbf{68.35} \\
            \textbf{Sinhala (si)} & 37.71 & 40.17 & 42.77 & 45.72 & \textbf{47.54} & 47.06 \\
            \textbf{Icelandic (is)} & 51.32 & 48.53 & 50.16 & 53.39 & \textbf{55.53} & 54.61 \\
            \textbf{Sindhi (sd)} & 27.28 & 26.46 & 31.08 & 32.42 & 35.05 & \textbf{35.28} \\
            \textbf{Basque (eu)} & 58.55 & 56.55 & 56.78 & 59.56 & 60.62 & \textbf{61.04} \\
            \textbf{Amharic (am)} & 24.10 & 28.57 & 35.01 & 34.20 & \textbf{37.47} & 36.87 \\
            \textbf{Lithuanian (lt)} & 71.31 & 69.50 & 69.65 & 72.26 & 72.43 & \textbf{73.09} \\
            \textbf{Welsh (cy)} & 58.36 & 58.50 & 57.56 & 59.66 & 59.95 & \textbf{60.24} \\
            \textbf{Haitian Creole (ht)} & 68.13 & 67.05 & 67.97 & 70.70 & 71.33 & \textbf{71.41} \\
            \midrule
            \textbf{Average} & 50.05 & 50.86 & 51.74 & 53.66 & 54.62 & \textbf{55.23} \\
            \bottomrule
        \end{tabular}
        }
    \end{minipage}
\end{table*}

\section{Related Works}
\smallsection{Pre-trained Language Models} 
Since their introduction~\cite{peters2018elmo, devlin2019bert}, pre-trained language models have enjoyed tremendous success in all aspects of natural language processing.
Follow-up research has advanced PLMs by further scaling them with respect to the number of parameters and training data.
PLM models have grown considerably in their sizes, from millions~\cite{devlin2019bert,yang2019xlnet} of parameters to billions~\cite{brown2020gpt3,raffel2020t5,shoeybi2019megatron} and even trillion-level~\cite{fedus2021switch}.
Another avenue of improvement has been improving the training objectives used to train PLMs.
A broad spectrum of pre-training objectives have been explored with different levels of success, 
Notable examples include masked language modeling~\cite{devlin2019bert}, auto-regressive causal language modeling~\cite{yang2019xlnet}, model-based denoising~\cite{clark2020electra}, and corrective language modeling~\cite{meng2021cocolm}.
Despite these innovations in scaling and pre-training objectives, the vast majority of the work has focused on text-only training objectives applied to general domain corpora, e.g., Wikipedia and CommonCrawl.
In this paper, we deviate from most previous works by exploring PLM training using solely Twitter in-domain data and training our model based on both text-based and social-based objectives.

\smallsection{Tweet Language Models}
While a majority of PLMs are trained on general domain corpora, a few language models have been proposed specifically for Twitter and other social media platforms. BERTweet~\cite{nguyen2020bertweet} mirrors BERT training on 850 million English Tweets. TimeLMs~\cite{loureiro2022timelm} trains a set of RoBERTa~\cite{liu2019roberta} models for English Tweets on different time ranges. XLM-T~\cite{barbieri2022xlmt} continues the pre-training process from an XLM-R~\cite{conneau2019unsupervised} checkpoint on 198 million multilingual Tweets. These methods mostly replicate existing general domain PLM methods and simply substitute the training data with Tweets.
However, our approach utilizes additional social engagement signals to enhance the pre-trained Tweet representations. 

\smallsection{Enriching PLMs with Additional Information}
Several existing works use additional information for language model pre-training. ERNIE~\cite{zhang2019ernie} and K-BERT~\cite{liu2020kbert} inject entities and their relations from knowledge graphs to augment the pre-training corpus. OAG-BERT~\cite{liu2021oagbert} appends metadata of a document to its raw text, and designs objectives to jointly predict text and metadata.
These works focus on bringing metadata and knowledge by injecting training instances, while our work leverages the rich social engagements embedded in the social media platform for text relevance.
Recent work~\cite{yasunaga2022linkbert} has utilized document hyperlinks for LM pre-training, but does so with a simple three-way classification objective.

\smallsection{Network Embedding} Network embedding has emerged as a valuable tool for transferring information from relational data to other tasks~\cite{el2022graph}.
Early network embedding methods such as DeepWalk~\cite{perozzi2014deepwalk} and node2vec~\cite{grover2016node2vec} embed homogeneous graphs by performing random walks and applying SkipGram modeling.
With the introduction of heterogeneous information networks~\cite{sun2013mining} as a formalism to model rich multi-typed, multi-relational networks, many heterogeneous network embedding approaches were developed~\cite{chang2015heterogeneous,tang2015pte,xu2017embedding,chen2017task,dong2017metapath2vec}.
However, many of these techniques are difficult to scale to very large networks. 
In this work, we apply knowledge graph embeddings~\cite{wang2017knowledge,bordes2013translating,trouillon2016complex}, which have been shown to be both highly scalable and flexible enough to model multiple node and edge types.

\section{Conclusions}
In this work we introduce \our, a multilingual language model trained on a large Tweet corpus. Unlike previous BERT-style language models, \our is trained using two objectives: (1) a standard MLM pre-training objective and (2) a contrasting social objective. We perform a variety of downstream tasks using \our on Tweet data. Our experiments demonstrate that \our outperforms previously released language models on both semantic and social engagement prediction tasks. We release \our\footnote{\url{http://huggingface.co/Twitter/twhin-bert-base}}\footnote{\url{http://huggingface.co/Twitter/twhin-bert-large}} to the academic community to further research in social media NLP.

\bibliographystyle{ACM-Reference-Format}
\bibliography{main}


\begin{thebibliography}{54}


\ifx \showCODEN    \undefined \def \showCODEN     #1{\unskip}     \fi
\ifx \showDOI      \undefined \def \showDOI       #1{#1}\fi
\ifx \showISBNx    \undefined \def \showISBNx     #1{\unskip}     \fi
\ifx \showISBNxiii \undefined \def \showISBNxiii  #1{\unskip}     \fi
\ifx \showISSN     \undefined \def \showISSN      #1{\unskip}     \fi
\ifx \showLCCN     \undefined \def \showLCCN      #1{\unskip}     \fi
\ifx \shownote     \undefined \def \shownote      #1{#1}          \fi
\ifx \showarticletitle \undefined \def \showarticletitle #1{#1}   \fi
\ifx \showURL      \undefined \def \showURL       {\relax}        \fi
\providecommand\bibfield[2]{#2}
\providecommand\bibinfo[2]{#2}
\providecommand\natexlab[1]{#1}
\providecommand\showeprint[2][]{arXiv:#2}

\bibitem[Alharbi et~al\mbox{.}(2020)]%
        {asad}
\bibfield{author}{\bibinfo{person}{Basma Alharbi}, \bibinfo{person}{Hind
  Alamro}, \bibinfo{person}{Manal~Abdulaziz Alshehri}, \bibinfo{person}{Zuhair
  Khayyat}, \bibinfo{person}{Manal Kalkatawi}, \bibinfo{person}{Inji~Ibrahim
  Jaber}, {and} \bibinfo{person}{Xiangliang Zhang}.}
  \bibinfo{year}{2020}\natexlab{}.
\newblock \showarticletitle{ASAD: A Twitter-based Benchmark Arabic Sentiment
  Analysis Dataset}.
\newblock \bibinfo{journal}{\emph{ArXiv}}  \bibinfo{volume}{abs/2011.00578}
  (\bibinfo{year}{2020}).
\newblock


\bibitem[Barbieri et~al\mbox{.}(2021)]%
        {barbieri2022xlmt}
\bibfield{author}{\bibinfo{person}{Francesco Barbieri},
  \bibinfo{person}{Luis~Espinosa Anke}, {and} \bibinfo{person}{Jos{\'e}
  Camacho-Collados}.} \bibinfo{year}{2021}\natexlab{}.
\newblock \showarticletitle{XLM-T: A Multilingual Language Model Toolkit for
  Twitter}.
\newblock \bibinfo{journal}{\emph{ArXiv}}  \bibinfo{volume}{abs/2104.12250}
  (\bibinfo{year}{2021}).
\newblock


\bibitem[Barbieri et~al\mbox{.}(2018)]%
        {semeval2018t2}
\bibfield{author}{\bibinfo{person}{Francesco Barbieri}, \bibinfo{person}{Jose
  Camacho-Collados}, \bibinfo{person}{Francesco Ronzano}, \bibinfo{person}{Luis
  Espinosa-Anke}, \bibinfo{person}{Miguel Ballesteros},
  \bibinfo{person}{Valerio Basile}, \bibinfo{person}{Viviana Patti}, {and}
  \bibinfo{person}{Horacio Saggion}.} \bibinfo{year}{2018}\natexlab{}.
\newblock \showarticletitle{{S}em{E}val 2018 Task 2: Multilingual Emoji
  Prediction}. In \bibinfo{booktitle}{\emph{Proceedings of The 12th
  International Workshop on Semantic Evaluation}}.
  \bibinfo{publisher}{Association for Computational Linguistics},
  \bibinfo{address}{New Orleans, Louisiana}, \bibinfo{pages}{24--33}.
\newblock
\urldef\tempurl%
\url{https://doi.org/10.18653/v1/S18-1003}
\showDOI{\tempurl}


\bibitem[Bojanowski et~al\mbox{.}(2017)]%
        {bojanowski2017fasttext}
\bibfield{author}{\bibinfo{person}{Piotr Bojanowski}, \bibinfo{person}{Edouard
  Grave}, \bibinfo{person}{Armand Joulin}, {and} \bibinfo{person}{Tomas
  Mikolov}.} \bibinfo{year}{2017}\natexlab{}.
\newblock \showarticletitle{Enriching Word Vectors with Subword Information}.
\newblock \bibinfo{journal}{\emph{Transactions of the Association for
  Computational Linguistics}}  \bibinfo{volume}{5} (\bibinfo{year}{2017}),
  \bibinfo{pages}{135--146}.
\newblock
\urldef\tempurl%
\url{https://doi.org/10.1162/tacl_a_00051}
\showDOI{\tempurl}


\bibitem[Bordes et~al\mbox{.}(2013)]%
        {bordes2013translating}
\bibfield{author}{\bibinfo{person}{Antoine Bordes}, \bibinfo{person}{Nicolas
  Usunier}, \bibinfo{person}{Alberto Garcia-Duran}, \bibinfo{person}{Jason
  Weston}, {and} \bibinfo{person}{Oksana Yakhnenko}.}
  \bibinfo{year}{2013}\natexlab{}.
\newblock \showarticletitle{Translating embeddings for modeling
  multi-relational data}.
\newblock \bibinfo{journal}{\emph{Advances in neural information processing
  systems}}  \bibinfo{volume}{26} (\bibinfo{year}{2013}).
\newblock


\bibitem[Brown et~al\mbox{.}(2020)]%
        {brown2020gpt3}
\bibfield{author}{\bibinfo{person}{Tom Brown}, \bibinfo{person}{Benjamin Mann},
  \bibinfo{person}{Nick Ryder}, \bibinfo{person}{Melanie Subbiah},
  \bibinfo{person}{Jared~D Kaplan}, \bibinfo{person}{Prafulla Dhariwal},
  \bibinfo{person}{Arvind Neelakantan}, \bibinfo{person}{Pranav Shyam},
  \bibinfo{person}{Girish Sastry}, \bibinfo{person}{Amanda Askell},
  {et~al\mbox{.}}} \bibinfo{year}{2020}\natexlab{}.
\newblock \showarticletitle{Language models are few-shot learners}.
\newblock \bibinfo{journal}{\emph{Advances in neural information processing
  systems}}  \bibinfo{volume}{33} (\bibinfo{year}{2020}),
  \bibinfo{pages}{1877--1901}.
\newblock


\bibitem[Chang et~al\mbox{.}(2015)]%
        {chang2015heterogeneous}
\bibfield{author}{\bibinfo{person}{S. Chang}, \bibinfo{person}{W. Han},
  \bibinfo{person}{J. Tang}, \bibinfo{person}{G. Qi}, \bibinfo{person}{C.
  Aggarwal}, {and} \bibinfo{person}{T. Huang}.}
  \bibinfo{year}{2015}\natexlab{}.
\newblock \showarticletitle{Heterogeneous network embedding via deep
  architectures}. In \bibinfo{booktitle}{\emph{SIGKDD}}.
  \bibinfo{pages}{119--128}.
\newblock


\bibitem[Chen et~al\mbox{.}(2020)]%
        {chen2020simclr}
\bibfield{author}{\bibinfo{person}{Ting Chen}, \bibinfo{person}{Simon
  Kornblith}, \bibinfo{person}{Mohammad Norouzi}, {and}
  \bibinfo{person}{Geoffrey Hinton}.} \bibinfo{year}{2020}\natexlab{}.
\newblock \showarticletitle{A Simple Framework for Contrastive Learning of
  Visual Representations}. In \bibinfo{booktitle}{\emph{Proceedings of the 37th
  International Conference on Machine Learning}}
  \emph{(\bibinfo{series}{Proceedings of Machine Learning Research},
  Vol.~\bibinfo{volume}{119})}, \bibfield{editor}{\bibinfo{person}{Hal~Daumé
  III} {and} \bibinfo{person}{Aarti Singh}} (Eds.). \bibinfo{publisher}{PMLR},
  \bibinfo{pages}{1597--1607}.
\newblock
\urldef\tempurl%
\url{https://proceedings.mlr.press/v119/chen20j.html}
\showURL{%
\tempurl}


\bibitem[Chen and Sun(2017)]%
        {chen2017task}
\bibfield{author}{\bibinfo{person}{T. Chen} {and} \bibinfo{person}{Y. Sun}.}
  \bibinfo{year}{2017}\natexlab{}.
\newblock \showarticletitle{Task-guided and path-augmented heterogeneous
  network embedding for author identification}. In
  \bibinfo{booktitle}{\emph{WSDM}}. \bibinfo{pages}{295--304}.
\newblock


\bibitem[Cheng et~al\mbox{.}(2014)]%
        {cheng2014cascade}
\bibfield{author}{\bibinfo{person}{Justin Cheng}, \bibinfo{person}{Lada~A.
  Adamic}, \bibinfo{person}{P.~Alex Dow}, \bibinfo{person}{Jon~M. Kleinberg},
  {and} \bibinfo{person}{Jure Leskovec}.} \bibinfo{year}{2014}\natexlab{}.
\newblock \showarticletitle{Can cascades be predicted?}. In
  \bibinfo{booktitle}{\emph{23rd International World Wide Web Conference, {WWW}
  '14, Seoul, Republic of Korea, April 7-11, 2014}},
  \bibfield{editor}{\bibinfo{person}{Chin{-}Wan Chung},
  \bibinfo{person}{Andrei~Z. Broder}, \bibinfo{person}{Kyuseok Shim}, {and}
  \bibinfo{person}{Torsten Suel}} (Eds.). \bibinfo{publisher}{{ACM}},
  \bibinfo{pages}{925--936}.
\newblock
\urldef\tempurl%
\url{https://doi.org/10.1145/2566486.2567997}
\showDOI{\tempurl}


\bibitem[Clark et~al\mbox{.}(2020)]%
        {clark2020electra}
\bibfield{author}{\bibinfo{person}{Kevin Clark}, \bibinfo{person}{Minh{-}Thang
  Luong}, \bibinfo{person}{Quoc~V. Le}, {and} \bibinfo{person}{Christopher~D.
  Manning}.} \bibinfo{year}{2020}\natexlab{}.
\newblock \showarticletitle{{ELECTRA:} Pre-training Text Encoders as
  Discriminators Rather Than Generators}. In \bibinfo{booktitle}{\emph{8th
  International Conference on Learning Representations, {ICLR} 2020, Addis
  Ababa, Ethiopia, April 26-30, 2020}}. \bibinfo{publisher}{OpenReview.net}.
\newblock
\urldef\tempurl%
\url{https://openreview.net/forum?id=r1xMH1BtvB}
\showURL{%
\tempurl}


\bibitem[Conneau et~al\mbox{.}(2020)]%
        {conneau2019unsupervised}
\bibfield{author}{\bibinfo{person}{Alexis Conneau}, \bibinfo{person}{Kartikay
  Khandelwal}, \bibinfo{person}{Naman Goyal}, \bibinfo{person}{Vishrav
  Chaudhary}, \bibinfo{person}{Guillaume Wenzek}, \bibinfo{person}{Francisco
  Guzm{\'a}n}, \bibinfo{person}{Edouard Grave}, \bibinfo{person}{Myle Ott},
  \bibinfo{person}{Luke Zettlemoyer}, {and} \bibinfo{person}{Veselin
  Stoyanov}.} \bibinfo{year}{2020}\natexlab{}.
\newblock \showarticletitle{Unsupervised Cross-lingual Representation Learning
  at Scale}. In \bibinfo{booktitle}{\emph{ACL}}.
\newblock


\bibitem[Conneau and Lample(2019)]%
        {conneau2019xlm}
\bibfield{author}{\bibinfo{person}{Alexis Conneau} {and}
  \bibinfo{person}{Guillaume Lample}.} \bibinfo{year}{2019}\natexlab{}.
\newblock \showarticletitle{Cross-lingual Language Model Pretraining}. In
  \bibinfo{booktitle}{\emph{Advances in Neural Information Processing Systems
  32: Annual Conference on Neural Information Processing Systems 2019, NeurIPS
  2019, December 8-14, 2019, Vancouver, BC, Canada}},
  \bibfield{editor}{\bibinfo{person}{Hanna~M. Wallach}, \bibinfo{person}{Hugo
  Larochelle}, \bibinfo{person}{Alina Beygelzimer}, \bibinfo{person}{Florence
  d'Alch{\'{e}}{-}Buc}, \bibinfo{person}{Emily~B. Fox}, {and}
  \bibinfo{person}{Roman Garnett}} (Eds.). \bibinfo{pages}{7057--7067}.
\newblock
\urldef\tempurl%
\url{https://proceedings.neurips.cc/paper/2019/hash/c04c19c2c2474dbf5f7ac4372c5b9af1-Abstract.html}
\showURL{%
\tempurl}


\bibitem[Devlin et~al\mbox{.}(2019)]%
        {devlin2019bert}
\bibfield{author}{\bibinfo{person}{Jacob Devlin}, \bibinfo{person}{Ming{-}Wei
  Chang}, \bibinfo{person}{Kenton Lee}, {and} \bibinfo{person}{Kristina
  Toutanova}.} \bibinfo{year}{2019}\natexlab{}.
\newblock \showarticletitle{{BERT:} Pre-training of Deep Bidirectional
  Transformers for Language Understanding}. In
  \bibinfo{booktitle}{\emph{Proceedings of the 2019 Conference of the North
  American Chapter of the Association for Computational Linguistics: Human
  Language Technologies, {NAACL-HLT} 2019, Minneapolis, MN, USA, June 2-7,
  2019, Volume 1 (Long and Short Papers)}},
  \bibfield{editor}{\bibinfo{person}{Jill Burstein}, \bibinfo{person}{Christy
  Doran}, {and} \bibinfo{person}{Thamar Solorio}} (Eds.).
  \bibinfo{publisher}{Association for Computational Linguistics},
  \bibinfo{pages}{4171--4186}.
\newblock
\urldef\tempurl%
\url{https://doi.org/10.18653/v1/n19-1423}
\showDOI{\tempurl}


\bibitem[Dong et~al\mbox{.}(2017)]%
        {dong2017metapath2vec}
\bibfield{author}{\bibinfo{person}{Y. Dong}, \bibinfo{person}{N. Chawla}, {and}
  \bibinfo{person}{A. Swami}.} \bibinfo{year}{2017}\natexlab{}.
\newblock \showarticletitle{metapath2vec: Scalable representation learning for
  heterogeneous networks}. In \bibinfo{booktitle}{\emph{SIGKDD}}.
  \bibinfo{pages}{135--144}.
\newblock


\bibitem[El-Kishky et~al\mbox{.}(2022a)]%
        {el2022graph}
\bibfield{author}{\bibinfo{person}{Ahmed El-Kishky}, \bibinfo{person}{Michael
  Bronstein}, \bibinfo{person}{Ying Xiao}, {and} \bibinfo{person}{Aria
  Haghighi}.} \bibinfo{year}{2022}\natexlab{a}.
\newblock \showarticletitle{Graph-based Representation Learning for Web-scale
  Recommender Systems}. In \bibinfo{booktitle}{\emph{Proceedings of the 28th
  ACM SIGKDD Conference on Knowledge Discovery and Data Mining}}.
  \bibinfo{pages}{4784--4785}.
\newblock


\bibitem[El-Kishky et~al\mbox{.}(2022b)]%
        {el2022knn}
\bibfield{author}{\bibinfo{person}{Ahmed El-Kishky}, \bibinfo{person}{Thomas
  Markovich}, \bibinfo{person}{Kenny Leung}, \bibinfo{person}{Frank Portman},
  {and} \bibinfo{person}{Aria Haghighi}.} \bibinfo{year}{2022}\natexlab{b}.
\newblock \showarticletitle{kNN-Embed: Locally Smoothed Embedding Mixtures For
  Multi-interest Candidate Retrieval}.
\newblock \bibinfo{journal}{\emph{arXiv preprint arXiv:2205.06205}}
  (\bibinfo{year}{2022}).
\newblock


\bibitem[El{-}Kishky et~al\mbox{.}(2022)]%
        {elkishky2022twhin}
\bibfield{author}{\bibinfo{person}{Ahmed El{-}Kishky}, \bibinfo{person}{Thomas
  Markovich}, \bibinfo{person}{Serim Park}, \bibinfo{person}{Chetan Verma},
  \bibinfo{person}{Baekjin Kim}, \bibinfo{person}{Ramy Eskander},
  \bibinfo{person}{Yury Malkov}, \bibinfo{person}{Frank Portman},
  \bibinfo{person}{Sof{\'{\i}}a Samaniego}, \bibinfo{person}{Ying Xiao}, {and}
  \bibinfo{person}{Aria Haghighi}.} \bibinfo{year}{2022}\natexlab{}.
\newblock \showarticletitle{TwHIN: Embedding the Twitter Heterogeneous
  Information Network for Personalized Recommendation}. In
  \bibinfo{booktitle}{\emph{{KDD} '22: The 28th {ACM} {SIGKDD} Conference on
  Knowledge Discovery and Data Mining, Washington, DC, USA, August 14 - 18,
  2022}}, \bibfield{editor}{\bibinfo{person}{Aidong Zhang} {and}
  \bibinfo{person}{Huzefa Rangwala}} (Eds.). \bibinfo{publisher}{{ACM}},
  \bibinfo{pages}{2842--2850}.
\newblock
\urldef\tempurl%
\url{https://doi.org/10.1145/3534678.3539080}
\showDOI{\tempurl}


\bibitem[Fedus et~al\mbox{.}(2021)]%
        {fedus2021switch}
\bibfield{author}{\bibinfo{person}{William Fedus}, \bibinfo{person}{Barret
  Zoph}, {and} \bibinfo{person}{Noam Shazeer}.}
  \bibinfo{year}{2021}\natexlab{}.
\newblock \showarticletitle{Switch Transformers: Scaling to Trillion Parameter
  Models with Simple and Efficient Sparsity}.
\newblock \bibinfo{journal}{\emph{CoRR}}  \bibinfo{volume}{abs/2101.03961}
  (\bibinfo{year}{2021}).
\newblock
\showeprint[arXiv]{2101.03961}
\urldef\tempurl%
\url{https://arxiv.org/abs/2101.03961}
\showURL{%
\tempurl}


\bibitem[Goldberg and Levy(2014)]%
        {goldberg2014word2vec}
\bibfield{author}{\bibinfo{person}{Y. Goldberg} {and} \bibinfo{person}{O.
  Levy}.} \bibinfo{year}{2014}\natexlab{}.
\newblock \showarticletitle{word2vec Explained: deriving Mikolov et al.'s
  negative-sampling word-embedding method}.
\newblock \bibinfo{journal}{\emph{arXiv preprint arXiv:1402.3722}}
  (\bibinfo{year}{2014}).
\newblock


\bibitem[Grover and Leskovec(2016)]%
        {grover2016node2vec}
\bibfield{author}{\bibinfo{person}{A. Grover} {and} \bibinfo{person}{J.
  Leskovec}.} \bibinfo{year}{2016}\natexlab{}.
\newblock \showarticletitle{node2vec: Scalable feature learning for networks}.
  In \bibinfo{booktitle}{\emph{SIGKDD}}. \bibinfo{pages}{855--864}.
\newblock


\bibitem[Jegou et~al\mbox{.}(2010)]%
        {jegou2010product}
\bibfield{author}{\bibinfo{person}{Herve Jegou}, \bibinfo{person}{Matthijs
  Douze}, {and} \bibinfo{person}{Cordelia Schmid}.}
  \bibinfo{year}{2010}\natexlab{}.
\newblock \showarticletitle{Product quantization for nearest neighbor search}.
\newblock \bibinfo{journal}{\emph{IEEE transactions on pattern analysis and
  machine intelligence}} \bibinfo{volume}{33}, \bibinfo{number}{1}
  (\bibinfo{year}{2010}), \bibinfo{pages}{117--128}.
\newblock


\bibitem[Johnson et~al\mbox{.}(2019)]%
        {johnson2019billion}
\bibfield{author}{\bibinfo{person}{Jeff Johnson}, \bibinfo{person}{Matthijs
  Douze}, {and} \bibinfo{person}{Herv{\'e} J{\'e}gou}.}
  \bibinfo{year}{2019}\natexlab{}.
\newblock \showarticletitle{Billion-scale similarity search with gpus}.
\newblock \bibinfo{journal}{\emph{IEEE Transactions on Big Data}}
  \bibinfo{volume}{7}, \bibinfo{number}{3} (\bibinfo{year}{2019}),
  \bibinfo{pages}{535--547}.
\newblock


\bibitem[Lerer et~al\mbox{.}(2019)]%
        {lerer2019pytorch}
\bibfield{author}{\bibinfo{person}{Adam Lerer}, \bibinfo{person}{Ledell Wu},
  \bibinfo{person}{Jiajun Shen}, \bibinfo{person}{Timothee Lacroix},
  \bibinfo{person}{Luca Wehrstedt}, \bibinfo{person}{Abhijit Bose}, {and}
  \bibinfo{person}{Alex Peysakhovich}.} \bibinfo{year}{2019}\natexlab{}.
\newblock \showarticletitle{Pytorch-biggraph: A large-scale graph embedding
  system}.
\newblock \bibinfo{journal}{\emph{arXiv preprint arXiv:1903.12287}}
  (\bibinfo{year}{2019}).
\newblock


\bibitem[Liu et~al\mbox{.}(2020)]%
        {liu2020kbert}
\bibfield{author}{\bibinfo{person}{Weijie Liu}, \bibinfo{person}{Peng Zhou},
  \bibinfo{person}{Zhe Zhao}, \bibinfo{person}{Zhiruo Wang},
  \bibinfo{person}{Qi Ju}, \bibinfo{person}{Haotang Deng}, {and}
  \bibinfo{person}{Ping Wang}.} \bibinfo{year}{2020}\natexlab{}.
\newblock \showarticletitle{K-BERT: Enabling Language Representation with
  Knowledge Graph}. In \bibinfo{booktitle}{\emph{AAAI}}.
\newblock


\bibitem[Liu et~al\mbox{.}(2022)]%
        {liu2021oagbert}
\bibfield{author}{\bibinfo{person}{Xiao Liu}, \bibinfo{person}{Da Yin},
  \bibinfo{person}{Jingnan Zheng}, \bibinfo{person}{Xingjian Zhang},
  \bibinfo{person}{P. Zhang}, \bibinfo{person}{Hongxia Yang},
  \bibinfo{person}{Yuxiao Dong}, {and} \bibinfo{person}{Jie Tang}.}
  \bibinfo{year}{2022}\natexlab{}.
\newblock \showarticletitle{OAG-BERT: Towards a Unified Backbone Language Model
  for Academic Knowledge Services}.
\newblock \bibinfo{journal}{\emph{Proceedings of the 28th ACM SIGKDD Conference
  on Knowledge Discovery and Data Mining}} (\bibinfo{year}{2022}).
\newblock


\bibitem[Liu et~al\mbox{.}(2019)]%
        {liu2019roberta}
\bibfield{author}{\bibinfo{person}{Yinhan Liu}, \bibinfo{person}{Myle Ott},
  \bibinfo{person}{Naman Goyal}, \bibinfo{person}{Jingfei Du},
  \bibinfo{person}{Mandar Joshi}, \bibinfo{person}{Danqi Chen},
  \bibinfo{person}{Omer Levy}, \bibinfo{person}{Mike Lewis},
  \bibinfo{person}{Luke Zettlemoyer}, {and} \bibinfo{person}{Veselin
  Stoyanov}.} \bibinfo{year}{2019}\natexlab{}.
\newblock \showarticletitle{RoBERTa: {A} Robustly Optimized {BERT} Pretraining
  Approach}.
\newblock \bibinfo{journal}{\emph{CoRR}}  \bibinfo{volume}{abs/1907.11692}
  (\bibinfo{year}{2019}).
\newblock
\showeprint[arXiv]{1907.11692}
\urldef\tempurl%
\url{http://arxiv.org/abs/1907.11692}
\showURL{%
\tempurl}


\bibitem[Loureiro et~al\mbox{.}(2022)]%
        {loureiro2022timelm}
\bibfield{author}{\bibinfo{person}{Daniel Loureiro}, \bibinfo{person}{Francesco
  Barbieri}, \bibinfo{person}{Leonardo Neves}, \bibinfo{person}{Luis~Espinosa
  Anke}, {and} \bibinfo{person}{Jos{\'{e}} Camacho{-}Collados}.}
  \bibinfo{year}{2022}\natexlab{}.
\newblock \showarticletitle{TimeLMs: Diachronic Language Models from Twitter}.
  In \bibinfo{booktitle}{\emph{Proceedings of the 60th Annual Meeting of the
  Association for Computational Linguistics, {ACL} 2022 - System
  Demonstrations, Dublin, Ireland, May 22-27, 2022}},
  \bibfield{editor}{\bibinfo{person}{Valerio Basile}, \bibinfo{person}{Zornitsa
  Kozareva}, {and} \bibinfo{person}{Sanja Stajner}} (Eds.).
  \bibinfo{publisher}{Association for Computational Linguistics},
  \bibinfo{pages}{251--260}.
\newblock
\urldef\tempurl%
\url{https://doi.org/10.18653/v1/2022.acl-demo.25}
\showDOI{\tempurl}


\bibitem[Meng et~al\mbox{.}(2021)]%
        {meng2021cocolm}
\bibfield{author}{\bibinfo{person}{Yu Meng}, \bibinfo{person}{Chenyan Xiong},
  \bibinfo{person}{Payal Bajaj}, \bibinfo{person}{Saurabh Tiwary},
  \bibinfo{person}{Paul Bennett}, \bibinfo{person}{Jiawei Han}, {and}
  \bibinfo{person}{Xia Song}.} \bibinfo{year}{2021}\natexlab{}.
\newblock \showarticletitle{{COCO-LM:} Correcting and Contrasting Text
  Sequences for Language Model Pretraining}. In
  \bibinfo{booktitle}{\emph{Advances in Neural Information Processing Systems
  34: Annual Conference on Neural Information Processing Systems 2021, NeurIPS
  2021, December 6-14, 2021, virtual}},
  \bibfield{editor}{\bibinfo{person}{Marc'Aurelio Ranzato},
  \bibinfo{person}{Alina Beygelzimer}, \bibinfo{person}{Yann~N. Dauphin},
  \bibinfo{person}{Percy Liang}, {and} \bibinfo{person}{Jennifer~Wortman
  Vaughan}} (Eds.). \bibinfo{pages}{23102--23114}.
\newblock
\urldef\tempurl%
\url{https://proceedings.neurips.cc/paper/2021/hash/c2c2a04512b35d13102459f8784f1a2d-Abstract.html}
\showURL{%
\tempurl}


\bibitem[Mikolov et~al\mbox{.}(2013)]%
        {mikolov2013distributed}
\bibfield{author}{\bibinfo{person}{T. Mikolov}, \bibinfo{person}{I. Sutskever},
  \bibinfo{person}{K. Chen}, \bibinfo{person}{G. Corrado}, {and}
  \bibinfo{person}{J. Dean}.} \bibinfo{year}{2013}\natexlab{}.
\newblock \showarticletitle{Distributed representations of words and phrases
  and their compositionality}.
\newblock \bibinfo{journal}{\emph{NeurIPS}}  \bibinfo{volume}{26}
  (\bibinfo{year}{2013}).
\newblock


\bibitem[Mireshghallah et~al\mbox{.}(2022)]%
        {mireshghallah2022}
\bibfield{author}{\bibinfo{person}{Fatemehsadat Mireshghallah},
  \bibinfo{person}{Nikolai Vogler}, \bibinfo{person}{Junxian He},
  \bibinfo{person}{Omar Florez}, \bibinfo{person}{Ahmed El-Kishky}, {and}
  \bibinfo{person}{Taylor Berg-Kirkpatrick}.} \bibinfo{year}{2022}\natexlab{}.
\newblock \showarticletitle{Non-Parametric Temporal Adaptation for Social Media
  Topic Classification}.
\newblock \bibinfo{journal}{\emph{arXiv preprint arXiv:2209.05706}}
  (\bibinfo{year}{2022}).
\newblock


\bibitem[Nguyen et~al\mbox{.}(2020)]%
        {nguyen2020bertweet}
\bibfield{author}{\bibinfo{person}{Dat~Quoc Nguyen}, \bibinfo{person}{Thanh
  Vu}, {and} \bibinfo{person}{Anh~Tuan Nguyen}.}
  \bibinfo{year}{2020}\natexlab{}.
\newblock \showarticletitle{BERTweet: {A} pre-trained language model for
  English Tweets}. In \bibinfo{booktitle}{\emph{Proceedings of the 2020
  Conference on Empirical Methods in Natural Language Processing: System
  Demonstrations, {EMNLP} 2020 - Demos, Online, November 16-20, 2020}},
  \bibfield{editor}{\bibinfo{person}{Qun Liu} {and} \bibinfo{person}{David
  Schlangen}} (Eds.). \bibinfo{publisher}{Association for Computational
  Linguistics}, \bibinfo{pages}{9--14}.
\newblock
\urldef\tempurl%
\url{https://doi.org/10.18653/v1/2020.emnlp-demos.2}
\showDOI{\tempurl}


\bibitem[Patwa et~al\mbox{.}(2020)]%
        {semeval2020t9}
\bibfield{author}{\bibinfo{person}{Parth Patwa}, \bibinfo{person}{Gustavo
  Aguilar}, \bibinfo{person}{Sudipta Kar}, \bibinfo{person}{Suraj Pandey},
  \bibinfo{person}{Srinivas PYKL}, \bibinfo{person}{Bj{\"{o}}rn Gamb{\"{a}}ck},
  \bibinfo{person}{Tanmoy Chakraborty}, \bibinfo{person}{Thamar Solorio}, {and}
  \bibinfo{person}{Amitava Das}.} \bibinfo{year}{2020}\natexlab{}.
\newblock \showarticletitle{SemEval-2020 Task 9: Overview of Sentiment Analysis
  of Code-Mixed Tweets}. In \bibinfo{booktitle}{\emph{Proceedings of the
  Fourteenth Workshop on Semantic Evaluation, SemEval@COLING 2020, Barcelona
  (online), December 12-13, 2020}},
  \bibfield{editor}{\bibinfo{person}{Aur{\'{e}}lie Herbelot},
  \bibinfo{person}{Xiaodan Zhu}, \bibinfo{person}{Alexis Palmer},
  \bibinfo{person}{Nathan Schneider}, \bibinfo{person}{Jonathan May}, {and}
  \bibinfo{person}{Ekaterina Shutova}} (Eds.).
  \bibinfo{publisher}{International Committee for Computational Linguistics},
  \bibinfo{pages}{774--790}.
\newblock
\urldef\tempurl%
\url{https://doi.org/10.18653/v1/2020.semeval-1.100}
\showDOI{\tempurl}


\bibitem[Perozzi et~al\mbox{.}(2014)]%
        {perozzi2014deepwalk}
\bibfield{author}{\bibinfo{person}{B. Perozzi}, \bibinfo{person}{R. Al-Rfou},
  {and} \bibinfo{person}{S. Skiena}.} \bibinfo{year}{2014}\natexlab{}.
\newblock \showarticletitle{Deepwalk: Online learning of social
  representations}. In \bibinfo{booktitle}{\emph{SIGKDD}}.
  \bibinfo{pages}{701--710}.
\newblock


\bibitem[Peters et~al\mbox{.}(2018)]%
        {peters2018elmo}
\bibfield{author}{\bibinfo{person}{Matthew~E. Peters}, \bibinfo{person}{Mark
  Neumann}, \bibinfo{person}{Mohit Iyyer}, \bibinfo{person}{Matt Gardner},
  \bibinfo{person}{Christopher Clark}, \bibinfo{person}{Kenton Lee}, {and}
  \bibinfo{person}{Luke Zettlemoyer}.} \bibinfo{year}{2018}\natexlab{}.
\newblock \showarticletitle{Deep Contextualized Word Representations}. In
  \bibinfo{booktitle}{\emph{Proceedings of the 2018 Conference of the North
  American Chapter of the Association for Computational Linguistics: Human
  Language Technologies, {NAACL-HLT} 2018, New Orleans, Louisiana, USA, June
  1-6, 2018, Volume 1 (Long Papers)}},
  \bibfield{editor}{\bibinfo{person}{Marilyn~A. Walker}, \bibinfo{person}{Heng
  Ji}, {and} \bibinfo{person}{Amanda Stent}} (Eds.).
  \bibinfo{publisher}{Association for Computational Linguistics},
  \bibinfo{pages}{2227--2237}.
\newblock
\urldef\tempurl%
\url{https://doi.org/10.18653/v1/n18-1202}
\showDOI{\tempurl}


\bibitem[Radford et~al\mbox{.}(2019)]%
        {radford2019gpt2}
\bibfield{author}{\bibinfo{person}{Alec Radford}, \bibinfo{person}{Jeffrey Wu},
  \bibinfo{person}{Rewon Child}, \bibinfo{person}{David Luan},
  \bibinfo{person}{Dario Amodei}, \bibinfo{person}{Ilya Sutskever},
  {et~al\mbox{.}}} \bibinfo{year}{2019}\natexlab{}.
\newblock \showarticletitle{Language models are unsupervised multitask
  learners}.
\newblock \bibinfo{journal}{\emph{OpenAI blog}} \bibinfo{volume}{1},
  \bibinfo{number}{8} (\bibinfo{year}{2019}), \bibinfo{pages}{9}.
\newblock


\bibitem[Raffel et~al\mbox{.}(2020)]%
        {raffel2020t5}
\bibfield{author}{\bibinfo{person}{Colin Raffel}, \bibinfo{person}{Noam
  Shazeer}, \bibinfo{person}{Adam Roberts}, \bibinfo{person}{Katherine Lee},
  \bibinfo{person}{Sharan Narang}, \bibinfo{person}{Michael Matena},
  \bibinfo{person}{Yanqi Zhou}, \bibinfo{person}{Wei Li}, {and}
  \bibinfo{person}{Peter~J. Liu}.} \bibinfo{year}{2020}\natexlab{}.
\newblock \showarticletitle{Exploring the Limits of Transfer Learning with a
  Unified Text-to-Text Transformer}.
\newblock \bibinfo{journal}{\emph{J. Mach. Learn. Res.}}  \bibinfo{volume}{21}
  (\bibinfo{year}{2020}), \bibinfo{pages}{140:1--140:67}.
\newblock
\urldef\tempurl%
\url{http://jmlr.org/papers/v21/20-074.html}
\showURL{%
\tempurl}


\bibitem[Rosenthal et~al\mbox{.}(2019)]%
        {semeval2017}
\bibfield{author}{\bibinfo{person}{Sara Rosenthal}, \bibinfo{person}{Noura
  Farra}, {and} \bibinfo{person}{Preslav Nakov}.}
  \bibinfo{year}{2019}\natexlab{}.
\newblock \showarticletitle{SemEval-2017 Task 4: Sentiment Analysis in
  Twitter}.
\newblock \bibinfo{journal}{\emph{CoRR}}  \bibinfo{volume}{abs/1912.00741}
  (\bibinfo{year}{2019}).
\newblock
\showeprint[arXiv]{1912.00741}
\urldef\tempurl%
\url{http://arxiv.org/abs/1912.00741}
\showURL{%
\tempurl}


\bibitem[Rust et~al\mbox{.}(2021)]%
        {rust2021multilingual}
\bibfield{author}{\bibinfo{person}{Phillip Rust}, \bibinfo{person}{Jonas
  Pfeiffer}, \bibinfo{person}{Ivan Vulic}, \bibinfo{person}{Sebastian Ruder},
  {and} \bibinfo{person}{Iryna Gurevych}.} \bibinfo{year}{2021}\natexlab{}.
\newblock \showarticletitle{How Good is Your Tokenizer? On the Monolingual
  Performance of Multilingual Language Models}. In
  \bibinfo{booktitle}{\emph{Proceedings of the 59th Annual Meeting of the
  Association for Computational Linguistics and the 11th International Joint
  Conference on Natural Language Processing, {ACL/IJCNLP} 2021, (Volume 1: Long
  Papers), Virtual Event, August 1-6, 2021}},
  \bibfield{editor}{\bibinfo{person}{Chengqing Zong}, \bibinfo{person}{Fei
  Xia}, \bibinfo{person}{Wenjie Li}, {and} \bibinfo{person}{Roberto Navigli}}
  (Eds.). \bibinfo{publisher}{Association for Computational Linguistics},
  \bibinfo{pages}{3118--3135}.
\newblock
\urldef\tempurl%
\url{https://doi.org/10.18653/v1/2021.acl-long.243}
\showDOI{\tempurl}


\bibitem[Sankar et~al\mbox{.}(2020)]%
        {sankar2020infvae}
\bibfield{author}{\bibinfo{person}{Aravind Sankar}, \bibinfo{person}{Xinyang
  Zhang}, \bibinfo{person}{Adit Krishnan}, {and} \bibinfo{person}{Jiawei Han}.}
  \bibinfo{year}{2020}\natexlab{}.
\newblock \showarticletitle{Inf-VAE: {A} Variational Autoencoder Framework to
  Integrate Homophily and Influence in Diffusion Prediction}. In
  \bibinfo{booktitle}{\emph{{WSDM} '20: The Thirteenth {ACM} International
  Conference on Web Search and Data Mining, Houston, TX, USA, February 3-7,
  2020}}, \bibfield{editor}{\bibinfo{person}{James Caverlee},
  \bibinfo{person}{Xia~(Ben) Hu}, \bibinfo{person}{Mounia Lalmas}, {and}
  \bibinfo{person}{Wei Wang}} (Eds.). \bibinfo{publisher}{{ACM}},
  \bibinfo{pages}{510--518}.
\newblock
\urldef\tempurl%
\url{https://doi.org/10.1145/3336191.3371811}
\showDOI{\tempurl}


\bibitem[Shoeybi et~al\mbox{.}(2019)]%
        {shoeybi2019megatron}
\bibfield{author}{\bibinfo{person}{Mohammad Shoeybi}, \bibinfo{person}{Mostofa
  Patwary}, \bibinfo{person}{Raul Puri}, \bibinfo{person}{Patrick LeGresley},
  \bibinfo{person}{Jared Casper}, {and} \bibinfo{person}{Bryan Catanzaro}.}
  \bibinfo{year}{2019}\natexlab{}.
\newblock \showarticletitle{Megatron-LM: Training Multi-Billion Parameter
  Language Models Using Model Parallelism}.
\newblock \bibinfo{journal}{\emph{CoRR}}  \bibinfo{volume}{abs/1909.08053}
  (\bibinfo{year}{2019}).
\newblock
\showeprint[arXiv]{1909.08053}
\urldef\tempurl%
\url{http://arxiv.org/abs/1909.08053}
\showURL{%
\tempurl}


\bibitem[Sun and Han(2013)]%
        {sun2013mining}
\bibfield{author}{\bibinfo{person}{Y. Sun} {and} \bibinfo{person}{J. Han}.}
  \bibinfo{year}{2013}\natexlab{}.
\newblock \showarticletitle{Mining heterogeneous information networks: a
  structural analysis approach}.
\newblock \bibinfo{journal}{\emph{Acm Sigkdd Explorations Newsletter}}
  (\bibinfo{year}{2013}).
\newblock


\bibitem[Suzuki(2019)]%
        {covid_jp}
\bibfield{author}{\bibinfo{person}{Yu Suzuki}.}
  \bibinfo{year}{2019}\natexlab{}.
\newblock \showarticletitle{Filtering Method for Twitter Streaming Data Using
  Human-in-the-Loop Machine Learning}.
\newblock \bibinfo{journal}{\emph{J. Inf. Process.}}  \bibinfo{volume}{27}
  (\bibinfo{year}{2019}), \bibinfo{pages}{404--410}.
\newblock
\urldef\tempurl%
\url{https://doi.org/10.2197/ipsjjip.27.404}
\showDOI{\tempurl}


\bibitem[Tang et~al\mbox{.}(2015a)]%
        {tang2015pte}
\bibfield{author}{\bibinfo{person}{J. Tang}, \bibinfo{person}{M. Qu}, {and}
  \bibinfo{person}{Q. Mei}.} \bibinfo{year}{2015}\natexlab{a}.
\newblock \showarticletitle{Pte: Predictive text embedding through large-scale
  heterogeneous text networks}. In \bibinfo{booktitle}{\emph{SIGKDD}}.
  \bibinfo{pages}{1165--1174}.
\newblock


\bibitem[Tang et~al\mbox{.}(2015b)]%
        {tang2015line}
\bibfield{author}{\bibinfo{person}{Jian Tang}, \bibinfo{person}{Meng Qu},
  \bibinfo{person}{Mingzhe Wang}, \bibinfo{person}{Ming Zhang},
  \bibinfo{person}{Jun Yan}, {and} \bibinfo{person}{Qiaozhu Mei}.}
  \bibinfo{year}{2015}\natexlab{b}.
\newblock \showarticletitle{{LINE:} Large-scale Information Network Embedding}.
  In \bibinfo{booktitle}{\emph{Proceedings of the 24th International Conference
  on World Wide Web, {WWW} 2015, Florence, Italy, May 18-22, 2015}},
  \bibfield{editor}{\bibinfo{person}{Aldo Gangemi}, \bibinfo{person}{Stefano
  Leonardi}, {and} \bibinfo{person}{Alessandro Panconesi}} (Eds.).
  \bibinfo{publisher}{{ACM}}, \bibinfo{pages}{1067--1077}.
\newblock
\urldef\tempurl%
\url{https://doi.org/10.1145/2736277.2741093}
\showDOI{\tempurl}


\bibitem[Trouillon et~al\mbox{.}(2016)]%
        {trouillon2016complex}
\bibfield{author}{\bibinfo{person}{T. Trouillon}, \bibinfo{person}{J Welbl},
  \bibinfo{person}{S. Riedel}, \bibinfo{person}{{\'E}. Gaussier}, {and}
  \bibinfo{person}{G. Bouchard}.} \bibinfo{year}{2016}\natexlab{}.
\newblock \showarticletitle{Complex embeddings for simple link prediction}. In
  \bibinfo{booktitle}{\emph{ICML}}. PMLR, \bibinfo{pages}{2071--2080}.
\newblock


\bibitem[Vaswani et~al\mbox{.}(2017)]%
        {vaswani2017attention}
\bibfield{author}{\bibinfo{person}{Ashish Vaswani}, \bibinfo{person}{Noam
  Shazeer}, \bibinfo{person}{Niki Parmar}, \bibinfo{person}{Jakob Uszkoreit},
  \bibinfo{person}{Llion Jones}, \bibinfo{person}{Aidan~N Gomez},
  \bibinfo{person}{{\L}ukasz Kaiser}, {and} \bibinfo{person}{Illia
  Polosukhin}.} \bibinfo{year}{2017}\natexlab{}.
\newblock \showarticletitle{Attention is all you need}.
\newblock \bibinfo{journal}{\emph{Advances in neural information processing
  systems}}  \bibinfo{volume}{30} (\bibinfo{year}{2017}).
\newblock


\bibitem[Wang et~al\mbox{.}(2017)]%
        {wang2017knowledge}
\bibfield{author}{\bibinfo{person}{Q. Wang}, \bibinfo{person}{Z. Mao},
  \bibinfo{person}{B. Wang}, {and} \bibinfo{person}{L. Guo}.}
  \bibinfo{year}{2017}\natexlab{}.
\newblock \showarticletitle{Knowledge graph embedding: A survey of approaches
  and applications}.
\newblock \bibinfo{journal}{\emph{TKDE}} \bibinfo{volume}{29},
  \bibinfo{number}{12} (\bibinfo{year}{2017}), \bibinfo{pages}{2724--2743}.
\newblock


\bibitem[Xu et~al\mbox{.}(2017)]%
        {xu2017embedding}
\bibfield{author}{\bibinfo{person}{L. Xu}, \bibinfo{person}{X. Wei},
  \bibinfo{person}{J. Cao}, {and} \bibinfo{person}{P. Yu}.}
  \bibinfo{year}{2017}\natexlab{}.
\newblock \showarticletitle{Embedding of embedding (EOE) joint embedding for
  coupled heterogeneous networks}. In \bibinfo{booktitle}{\emph{WSDM}}.
  \bibinfo{pages}{741--749}.
\newblock


\bibitem[Xue et~al\mbox{.}(2021)]%
        {xue2021mt5}
\bibfield{author}{\bibinfo{person}{Linting Xue}, \bibinfo{person}{Noah
  Constant}, \bibinfo{person}{Adam Roberts}, \bibinfo{person}{Mihir Kale},
  \bibinfo{person}{Rami Al{-}Rfou}, \bibinfo{person}{Aditya Siddhant},
  \bibinfo{person}{Aditya Barua}, {and} \bibinfo{person}{Colin Raffel}.}
  \bibinfo{year}{2021}\natexlab{}.
\newblock \showarticletitle{mT5: {A} Massively Multilingual Pre-trained
  Text-to-Text Transformer}. In \bibinfo{booktitle}{\emph{Proceedings of the
  2021 Conference of the North American Chapter of the Association for
  Computational Linguistics: Human Language Technologies, {NAACL-HLT} 2021,
  Online, June 6-11, 2021}}, \bibfield{editor}{\bibinfo{person}{Kristina
  Toutanova}, \bibinfo{person}{Anna Rumshisky}, \bibinfo{person}{Luke
  Zettlemoyer}, \bibinfo{person}{Dilek Hakkani{-}T{\"{u}}r},
  \bibinfo{person}{Iz~Beltagy}, \bibinfo{person}{Steven Bethard},
  \bibinfo{person}{Ryan Cotterell}, \bibinfo{person}{Tanmoy Chakraborty}, {and}
  \bibinfo{person}{Yichao Zhou}} (Eds.). \bibinfo{publisher}{Association for
  Computational Linguistics}, \bibinfo{pages}{483--498}.
\newblock
\urldef\tempurl%
\url{https://doi.org/10.18653/v1/2021.naacl-main.41}
\showDOI{\tempurl}


\bibitem[Yang et~al\mbox{.}(2019)]%
        {yang2019xlnet}
\bibfield{author}{\bibinfo{person}{Zhilin Yang}, \bibinfo{person}{Zihang Dai},
  \bibinfo{person}{Yiming Yang}, \bibinfo{person}{Jaime~G. Carbonell},
  \bibinfo{person}{Ruslan Salakhutdinov}, {and} \bibinfo{person}{Quoc~V. Le}.}
  \bibinfo{year}{2019}\natexlab{}.
\newblock \showarticletitle{XLNet: Generalized Autoregressive Pretraining for
  Language Understanding}. In \bibinfo{booktitle}{\emph{Advances in Neural
  Information Processing Systems 32: Annual Conference on Neural Information
  Processing Systems 2019, NeurIPS 2019, December 8-14, 2019, Vancouver, BC,
  Canada}}, \bibfield{editor}{\bibinfo{person}{Hanna~M. Wallach},
  \bibinfo{person}{Hugo Larochelle}, \bibinfo{person}{Alina Beygelzimer},
  \bibinfo{person}{Florence d'Alch{\'{e}}{-}Buc}, \bibinfo{person}{Emily~B.
  Fox}, {and} \bibinfo{person}{Roman Garnett}} (Eds.).
  \bibinfo{pages}{5754--5764}.
\newblock
\urldef\tempurl%
\url{https://proceedings.neurips.cc/paper/2019/hash/dc6a7e655d7e5840e66733e9ee67cc69-Abstract.html}
\showURL{%
\tempurl}


\bibitem[Yasunaga et~al\mbox{.}(2022)]%
        {yasunaga2022linkbert}
\bibfield{author}{\bibinfo{person}{Michihiro Yasunaga}, \bibinfo{person}{Jure
  Leskovec}, {and} \bibinfo{person}{Percy Liang}.}
  \bibinfo{year}{2022}\natexlab{}.
\newblock \showarticletitle{{L}ink{BERT}: Pretraining Language Models with
  Document Links}. In \bibinfo{booktitle}{\emph{Proceedings of the 60th Annual
  Meeting of the Association for Computational Linguistics (Volume 1: Long
  Papers)}}. \bibinfo{publisher}{Association for Computational Linguistics},
  \bibinfo{address}{Dublin, Ireland}, \bibinfo{pages}{8003--8016}.
\newblock
\urldef\tempurl%
\url{https://doi.org/10.18653/v1/2022.acl-long.551}
\showDOI{\tempurl}


\bibitem[Ying et~al\mbox{.}(2018)]%
        {ying2018pinsage}
\bibfield{author}{\bibinfo{person}{Rex Ying}, \bibinfo{person}{Ruining He},
  \bibinfo{person}{Kaifeng Chen}, \bibinfo{person}{Pong Eksombatchai},
  \bibinfo{person}{William~L. Hamilton}, {and} \bibinfo{person}{Jure
  Leskovec}.} \bibinfo{year}{2018}\natexlab{}.
\newblock \showarticletitle{Graph Convolutional Neural Networks for Web-Scale
  Recommender Systems}. In \bibinfo{booktitle}{\emph{Proceedings of the 24th
  {ACM} {SIGKDD} International Conference on Knowledge Discovery {\&} Data
  Mining, {KDD} 2018, London, UK, August 19-23, 2018}},
  \bibfield{editor}{\bibinfo{person}{Yike Guo} {and} \bibinfo{person}{Faisal
  Farooq}} (Eds.). \bibinfo{publisher}{{ACM}}, \bibinfo{pages}{974--983}.
\newblock
\urldef\tempurl%
\url{https://doi.org/10.1145/3219819.3219890}
\showDOI{\tempurl}


\bibitem[Zhang et~al\mbox{.}(2019)]%
        {zhang2019ernie}
\bibfield{author}{\bibinfo{person}{Zhengyan Zhang}, \bibinfo{person}{Xu Han},
  \bibinfo{person}{Zhiyuan Liu}, \bibinfo{person}{Xin Jiang},
  \bibinfo{person}{Maosong Sun}, {and} \bibinfo{person}{Qun Liu}.}
  \bibinfo{year}{2019}\natexlab{}.
\newblock \showarticletitle{{ERNIE:} Enhanced Language Representation with
  Informative Entities}. In \bibinfo{booktitle}{\emph{Proceedings of the 57th
  Conference of the Association for Computational Linguistics, {ACL} 2019,
  Florence, Italy, July 28- August 2, 2019, Volume 1: Long Papers}},
  \bibfield{editor}{\bibinfo{person}{Anna Korhonen}, \bibinfo{person}{David~R.
  Traum}, {and} \bibinfo{person}{Llu{\'{\i}}s M{\`{a}}rquez}} (Eds.).
  \bibinfo{publisher}{Association for Computational Linguistics},
  \bibinfo{pages}{1441--1451}.
\newblock
\urldef\tempurl%
\url{https://doi.org/10.18653/v1/p19-1139}
\showDOI{\tempurl}


\end{thebibliography}

\appendix
\section{Distribution of Languages in Training Dataset}
\label{appendix:data_distribution}
\begin{figure*}[t]
    \centering
    \includegraphics[width=0.9\linewidth]{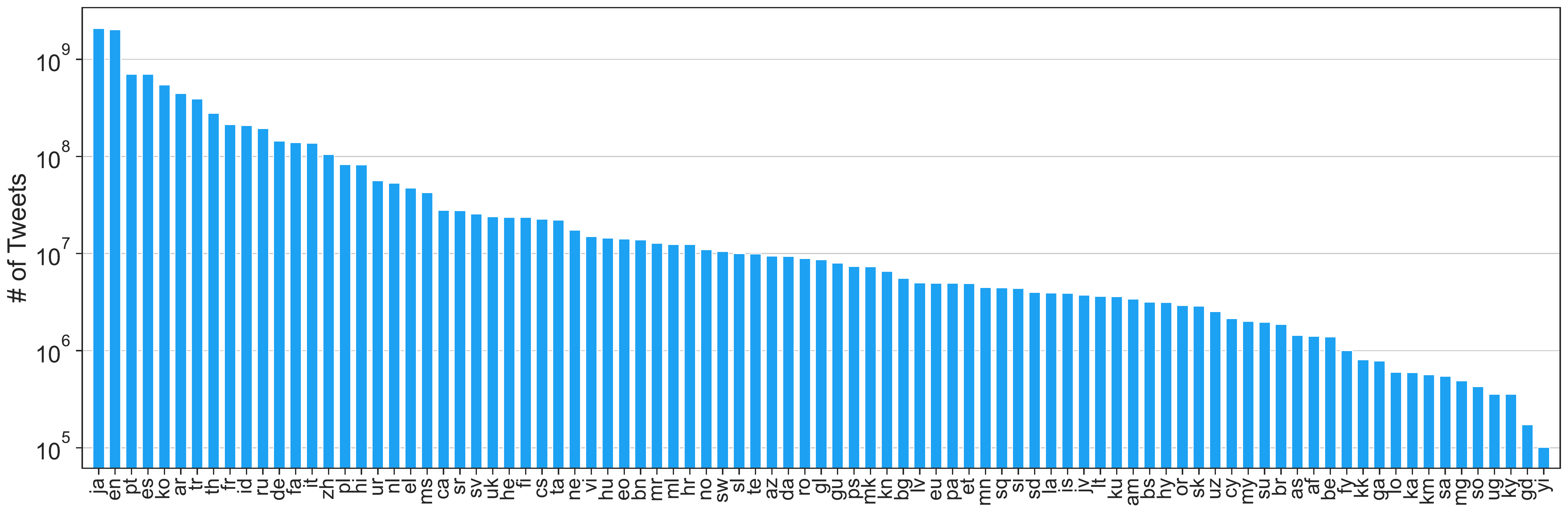}
    \caption{The number of Tweets in the pre-training dataset for each language. Languages are marked by ISO language codes. }
    \label{fig:lang_stats}
\end{figure*}

Figure~\ref{fig:lang_stats} shows the distribution of languages in our pre-training dataset.
Some languages with different variations (e.g., Hindi and Hindi Romanized) are represented with the same ISO language code.
We run fastText~\cite{bojanowski2017fasttext} language identification model \texttt{lid.176.bin}\footnote{\url{https://fasttext.cc/docs/en/language-identification.html}} to detect languages.

We deem a language ``high-resource'' if we have more than $10^8$ Tweets during pre-training \emph{after} frequency-based re-sampling (Section~\ref{sec:pretrain_setup}); ``mid-resource'' if we have more than $10^7$ and less than $10^8$ Tweets; ``low-resource'' if we have less than $10^7$ Tweets.

\section{Hyperparameters for Pre-training and Fine-Tuning}
\label{appendix:hyperparameters}
Table~\ref{tbl:hyperparameter_pretrain} shows the pre-training hyperparameters.
The model architecture and hyperparameters not shown in the table are the same as RoBERTa~\cite{liu2019roberta}.

Table~\ref{tbl:hyperparameter_fine-tune} shows the hyperparameters for classification fine-tuning.
We do hyperparameter selection on the development datasets and share the same set of hyperparameters for the base models, as we find them to perform well with this setting.
The weight decay for base models is set to zero.
A different set of hyperparameters were necessary for the large model because it behaves differently from the base models in terms of convergence.

\section{Evaluation Metrics for External Classification Benchmarks}
\label{appendix:metrics}
The recommended evaluation metrics that we report in Table~\ref{tbl:cls_results} are as follows.
Average recall for ASAD, SemEval 2017 datasets;
Macro-F1 for SemEval 2018 English and Spanish datasets;
Accuracy for COVID-JA, SemEval 2020 datasets.

\section{Engagement Prediction Results on Additional Languages}
\label{appendix:engagement}
Table~\ref{tbl:engagement_all_lang} shows the engagement prediction results on all available evaluation languages.
Some languages have more examples than other languages due to data availability.

\section{Hashtag Prediction Results on Additional Languages}
\label{appendix:hashtag}
Table~\ref{tbl:hashtag_all_lang} shows the hashtag prediction results on all available evaluation languages.
A small number of languages have less examples than shown in Table~\ref{tbl:text_classification_stats} due to data availability.
The Russian language is not evaluated as the XLM-T baseline fails on some Russian characters in our dataset.

\begin{table*}[t]
    \center
    \caption{Hyperparameters for pre-training TwHIN-BERT.}
    \label{tbl:hyperparameter_pretrain}
    \scalebox{0.95}{
    \begin{tabular}{l cc}
        \toprule
        \textbf{Hyperparameter} & \textbf{\our-base} & \textbf{\our-large} \\
        \midrule
        Max sequence length & 128 & 128 \\
        Precision & BF16 & BF16 \\
        \midrule
        \multicolumn{3}{l}{\textbf{Stage 1: MLM}} \\
        Total batch size & 6K & 8K \\
        Gradient accumulation steps & 1 & 4 \\
        Peak learning rate & 2e-4 & 2e-4 \\
        Warmup steps & 30K & 30K \\
        Total steps & 500K & 500K \\
        \midrule
        \multicolumn{3}{l}{\textbf{Stage 2: MLM + Social}} \\
        Total batch size & 6K & 6K \\
        Gradient checkpointing & No & Yes \\
        Peak learning rate & 1e-4 & 1e-4 \\
        Warmup steps & 30K & 30K \\
        Total steps & 500K & 500K \\
        Contrastive projection head & [768, 768] & [1024, 512] \\
        Contrastive loss temperature & 0.1 & 0.1 \\
        Loss balancing $\lambda$ & 0.05 & 0.05 \\
        \bottomrule
    \end{tabular}
    }
\end{table*}

\begin{table*}[t]
    \center
    \caption{Hyperparameters for fine-tuning \our and the baselines for classification.}
    \label{tbl:hyperparameter_fine-tune}
    \scalebox{0.95}{
    \begin{tabular}{l cccccc}
        \toprule
        \textbf{Hyperparameter} & \textbf{Hashtag} & \textbf{SE2017} & \textbf{SE2018} & \textbf{ASAD} & \textbf{COVID-JA} & \textbf{SE2020} \\
        \midrule
        \textbf{Base models} \\
        Learning rate & 4e-5 & 4e-5 & 1e-5 & 1e-5 & 2e-5 & 2e-5  \\
        Batch size & 128 & 128 & 128 & 128 & 128 & 128 \\
        \midrule
        \textbf{\our-large} \\
        Learning rate & 2e-5 & 2e-5 & 1e-5 & 1e-5 & 1e-5 & 1e-5 \\
        Weight decay & 0 & 0 & 5e-4 & 5e-4 & 0 & 5e-4 \\
        Batch size & 128 & 128 & 128 & 128 & 128 & 128  \\
        \bottomrule
    \end{tabular}
    }
\end{table*}

\end{document}